\documentclass[11pt]{article}

\usepackage[english]{babel}

\usepackage[letterpaper,top=2cm,bottom=2cm,left=3cm,right=3cm,marginparwidth=1.75cm]{geometry}

\usepackage{amsfonts,amssymb,amsmath,amscd}
\usepackage{amsmath}
\usepackage{floatrow}
\usepackage{pdfpages}
\usepackage{graphicx,wrapfig,lipsum}
\usepackage{amssymb}
\usepackage{mathrsfs}
\usepackage{latexsym}
\usepackage{enumerate}
\usepackage{listings}
\usepackage{color}
\usepackage{hyperref}
\usepackage{ulem}
\usepackage{sidecap}
\usepackage{amsfonts}
\usepackage{verbatim}
\usepackage{amsthm}
\usepackage{dsfont}
\usepackage{url}
\usepackage{tocloft}
\usepackage{float} 
\usepackage{ae}
\usepackage{xcolor}
\usepackage{amsfonts}
\usepackage{amsmath}
\usepackage{bm}
\usepackage{mathtools}
\usepackage{dsfont}
\usepackage{graphicx}
\usepackage{soul}
\usepackage{titlesec}
\usepackage[utf8]{inputenc}
\usepackage{amsthm}
  
\usepackage{sectsty}
\usepackage{enumitem}
\usepackage{titletoc}
\usepackage{xr}

\newtheorem{theorem}{Theorem}[section]

\numberwithin{equation}{section}

\title{The Double Descent Behavior in Two Layer Neural Network for Binary Classification}
\author{
Chathurika S Abeykoon\\
abeykoonc@rhodes.edu\\
  \and
Aleksandr Beknazaryan\\
beknazar@ucmail.uc.edu\\
  \and
Hailin Sang\\
sang@olemiss.edu
   }
\date{}
\begin{document}
\maketitle
\begin{abstract}
Recent studies observed a surprising concept on model test error called the double descent phenomenon where the increasing model complexity decreases the test error first and then the error increases and decreases again. To observe this, we work on a two-layer neural network model with a ReLU activation function ($\sigma(z)=\max(0,z)$) designed for binary classification under supervised learning. Our aim is to observe and investigate the mathematical theory behind the double descent behavior of model test error for varying model sizes. We quantify the model size by $\alpha=n/d$ where $n$ is the number of training samples and $d$ is the dimension of the model. Due to the complexity of the empirical risk minimization procedure, we use the Convex Gaussian MinMax Theorem to find a suitable candidate for the global training loss. \\

\end{abstract}


\section{Introduction}
Modern machine learning models are increasingly capable of mimicking human behavior, and are now commonly used in applications such as image and speech recognition, natural language processing, game playing, self-driving cars, and bioinformatics. A key component of these advancements is the neural network, a type of algorithm inspired by the human brain's neural networks. Neural networks have significantly advanced the field of machine learning, enabling more effective decision-making and task execution.

The underlying concepts of these applications use over-parameterized models where the model has more parameters than the data points in the training set. This refers not only to the number of parameters but also to the model's capacity to memorize data, where the number of parameters is one simple measure for it. The presence of more parameters not only makes these models to be complex but also generalizes well with the previously unseen data. The best model refers to the possible lowest ``test error" also known as the generalization error, which is a measure of how accurately the algorithm is able to predict outcome values for previously unseen data. 

For decades, the ``U-shaped curve'' explained the conventional wisdom of generalization error with respect to increasing model complexity where the generalization error decreases and increases again due to the bias-variance trade-off scenario. This classical wisdom in statistical learning focuses on finding the ``sweet spot" or the bottom of the U-shaped curve that refers to the lowest possible testing error balancing the under-fitting and over-fitting. This well-established idea focuses on controlling the model complexity to find the best fit. The recent studies by Belkin, Hsu, Ma, and Mandal in \cite{BHMM}, proposed a surprising behavior on generalization error in a prediction problem called the ``Double-Descent" phenomenon where a second descent happens after the classical U-shaped curve for increasing model complexity. 

\begin{wrapfigure}{r}{8cm}
\includegraphics[scale=0.35]{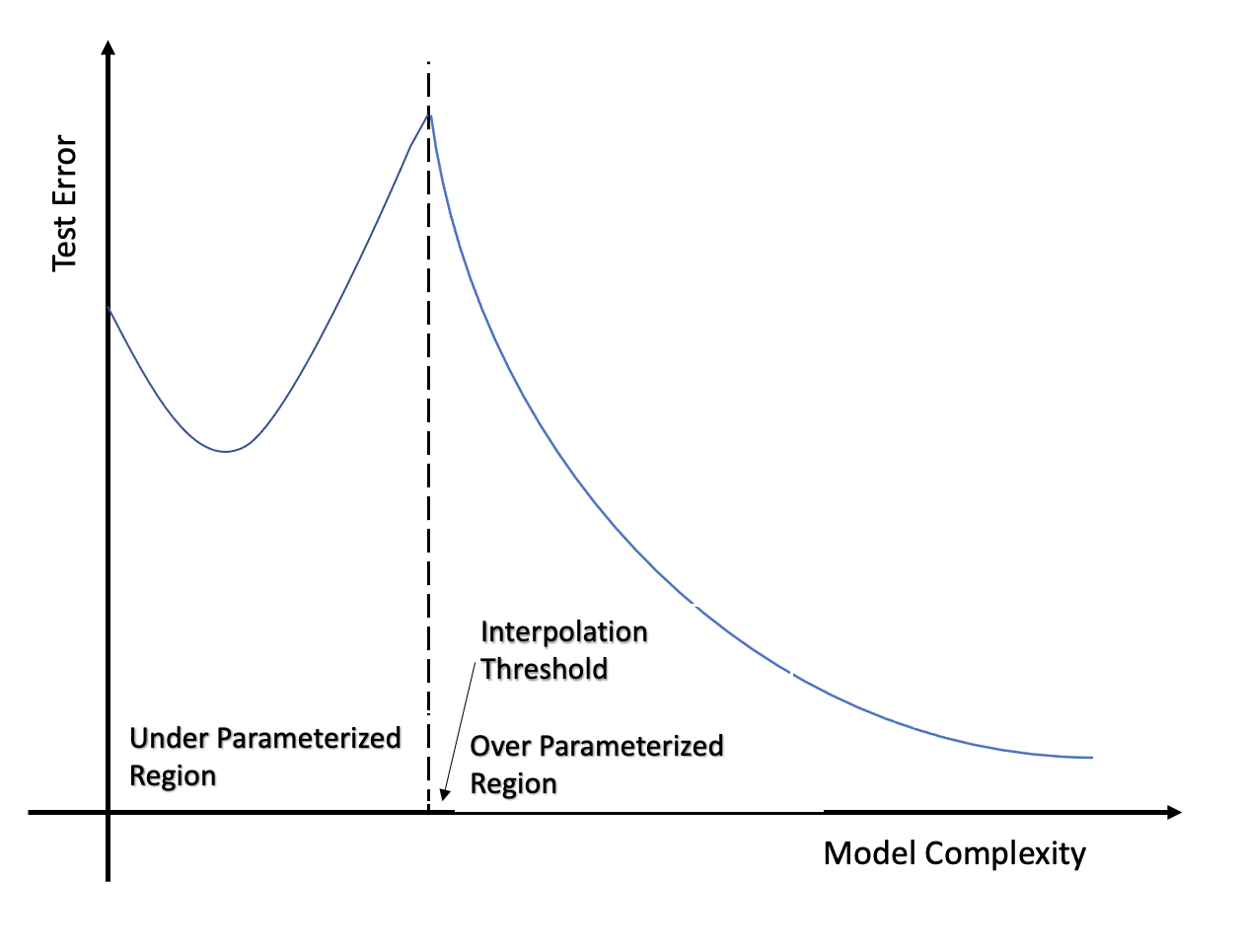}
\caption{\small{The double descent phenomenon in highly over-parameterized models.}}
\label{figdd}
\end{wrapfigure} 

With the double descent behavior, the test error first decreases and then increases tracing the U-shaped curve, and decreases again after the model complexity goes through a certain threshold value. The original concept in \cite{BHMM} analyzes the test error with respect to function class sizes or specifically by the number of parameters needed ($p$) and the number of training samples ($n$), so that around $p\approx n$, the model enters the modern interpolating regime. This transition threshold ($p\approx n$) is also known as the interpolation threshold which separates the under-parametrized region (the classical regime) and the over-parametrized region (modern interpolating regime). Belkin et al. in \cite{BHMM} empirically show the existence of this behavior in neural networks, decision trees and ensemble methods.

Large neural networks have the capacity to perform a second descent in over-parametrized regions due to their increased complexity. This has been demonstrated experimentally for many machine learning architectures like decision trees, two/multi-layer neural networks, and random features while some other studies like \cite{NKBYB+}, \cite{GJSGS+} and \cite{SGDSB+} explain this phenomenon related to ResNets, CNNs, and Transformers. Hence the empirical results of these models are more successful than the theoretical findings related to double descent. 

\subsection{Various forms of double descent}
Recent literature on double descent behavior studies the sensitivity of test error to the change in different settings/ variables in the model and demonstrates the presence of double descent.

The model-wise double descent is observed when the test error is expressed as a function of varying number of parameters $p$ while the model dimension $d$ and number of samples $n$ are fixed. This is the original double descent concept explained by Belkin et al. in \cite{BHMM}. The epoch-wise double descent analyzes the test error for varying training time while keeping the model dimension $d$ and the number of samples $n$ fixed. Higher the number of epochs, the longer the training time is needed. The longer we train, the better the performance. The learning rates, training procedures, noises in the model, and optimization methods may have considerable effect on the double descent curve based on epoch ( \cite{AAH}, \cite{NP} and \cite{NKBYB+}). In sample-wise double descent, the number of observations in the training procedure increases in order to observe the double descent behavior while the model dimension and the number of parameters are fixed.  This concept is studied experimentally in \cite{NKBYB+}, with reference to effective model complexity and they observe the peak when the number of samples matches the effective model complexity (\cite{NP}, \cite{NVKM})

A new approach to explain over-parameterization is to use the ratio between its training sample size and the number of parameters in the model. This ratio is often termed as the ``over-parameterization ratio'' where it enables us to decide the parameterization in two regions. The sensitivity of the test error for the ratios between $n, p$ (number of parameters), and $d$ are also observed both experimentally and theoretically in modern machine learning literature (\cite{KT}, \cite{DKT} and \cite{SGDSB+}).

\subsection{Our contribution}
Inspired by the fascinating studies on ratio-based double descent behavior, we work on the mathematical and statistical evaluation of a simple machine learning architecture for binary classification problem. Compared to the above studies, our work is new as follows.

We study the double descent of the test error in a binary linear classification problem using the student model: a two-layer neural network equipped with a ReLU activation function. The $n$ training data are generated from a teacher model which assigns each feature vector $\textbf{x}_i$ a binary label $y_i$ using two Gaussian vectors in $d$ dimension. We would call our work the ``ratio-wise double descent" and we quantify the over-parameterization by the ratio $\alpha=n/d$. We study the double descent behavior of the test error by treating it as a function of $\alpha$ when $n, d\to\infty$. The over-parameterization and under-parameterization regions can be easily separated as the following. 
\begin{itemize}
\item{$\alpha<1$ implies that $n<d$. This defines the over-parameterized region of our model.}
\item{$\alpha>1$ implies that $d<n$. This defines the under-parameterized region of our model.}
\item{$\alpha=1$ implies that $n=d$. This point separates the two regions as over and under-parameterized regions.}
\end{itemize}
As we increase $\alpha$, the model first goes through the over-parameterized region and then passes to the under-parameterized region. We identify this as another difference between the model-wise and ratio-wise double descents because, in the model-wise double descent, the test error switches from under-parameterized to over-parameterized regions when model complexity increases. 

We do not perform any training algorithm and instead, we use Convex Gaussian Min-max Theorem (CGMT) to find a theoretical candidate for the local minimum of the training risk. We observe that in the higher dimension, this candidate and global training loss have similar behaviors.
With respect to a specific loss function, 
the final test error when $n, d\to\infty $ is a function of $\alpha$ and we use this to view the double descent phenomena in our model for binary classification. Theorem \ref{The1} provide the generalization error formula of the student model in terms of parameters and Theorem \ref{mainT} give us values of these parameters for minimized empirical loss when $n, d\to\infty $ and they are valid for any margin-based loss function. We present the graphical results based on the square loss function and the two theorems at the end.

\section{Visualizing the double descent in binary classification models}\label{viz}
Before working on the theoretical analysis of the double descent occurrence in a simple binary classification model, below we visualize the double descent concept in the test error of the famous $\href{https://archive.ics.uci.edu/dataset/17/breast+cancer+wisconsin+diagnostic}{Wisconsin Breast Cancer}$ dataset. This dataset, widely used in classification tasks, contains 30 features such as tumor size and texture, which can be used to predict whether a tumor is malignant or benign. We vary the complexity of the binary classification model by increasing the sample size used to train the model while having the dimension fixed. We observe that the occurrence of a second descent is affected by factors like the regularization strength, the number of hidden nodes in the model, and the number of training epochs.

The following figures illustrate the behavior of test error as a function of $\alpha$ where $\alpha=n/d$ for 200 epochs in a two-layer binary classification neural network. The model applies ReLU activation function and it consists of one hidden layer with 10 neurons. The output layer has 2 neurons equipped with a sigmoid activation function for binary classification probabilities. It uses Adam optimizer for training and  binary cross-entropy as the loss function. For R code, see
supplementary material 2 S.1.

\begin{figure}[H]
\centering
{\caption{Test error showing the double descent behavior when a two-layer ReLu model is used with very low $l_2$ regularization $\lambda=10^{-6}$ for binary classification in Wisconsin Breast Cancer dataset. The test error decreases first and shows a slight increase around $\alpha=1$ and decreases again in the under-parameterized region.}\label{figa}}
{\includegraphics[scale=0.27]{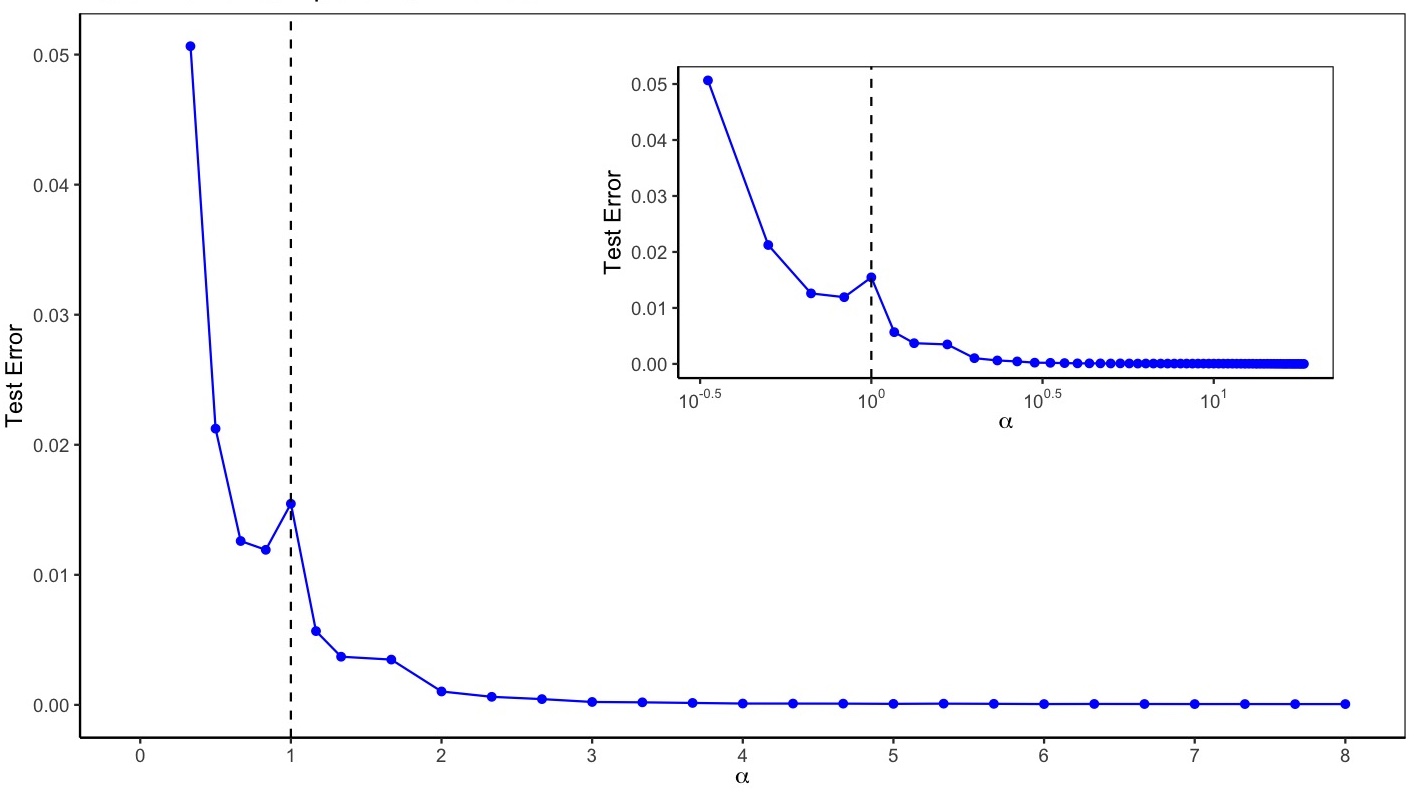}}
\end{figure}

\begin{figure}[H]
\centering
{\caption{Test error not showing the double descent behavior when a two-layer ReLu model is used with high $l_2$ regularization $\lambda=0.1$ for binary classification in Wisconsin Breast Cancer dataset. The test error decreases monotonically in both over and under parameterized regions.}\label{figb}}
{\includegraphics[scale=0.27]{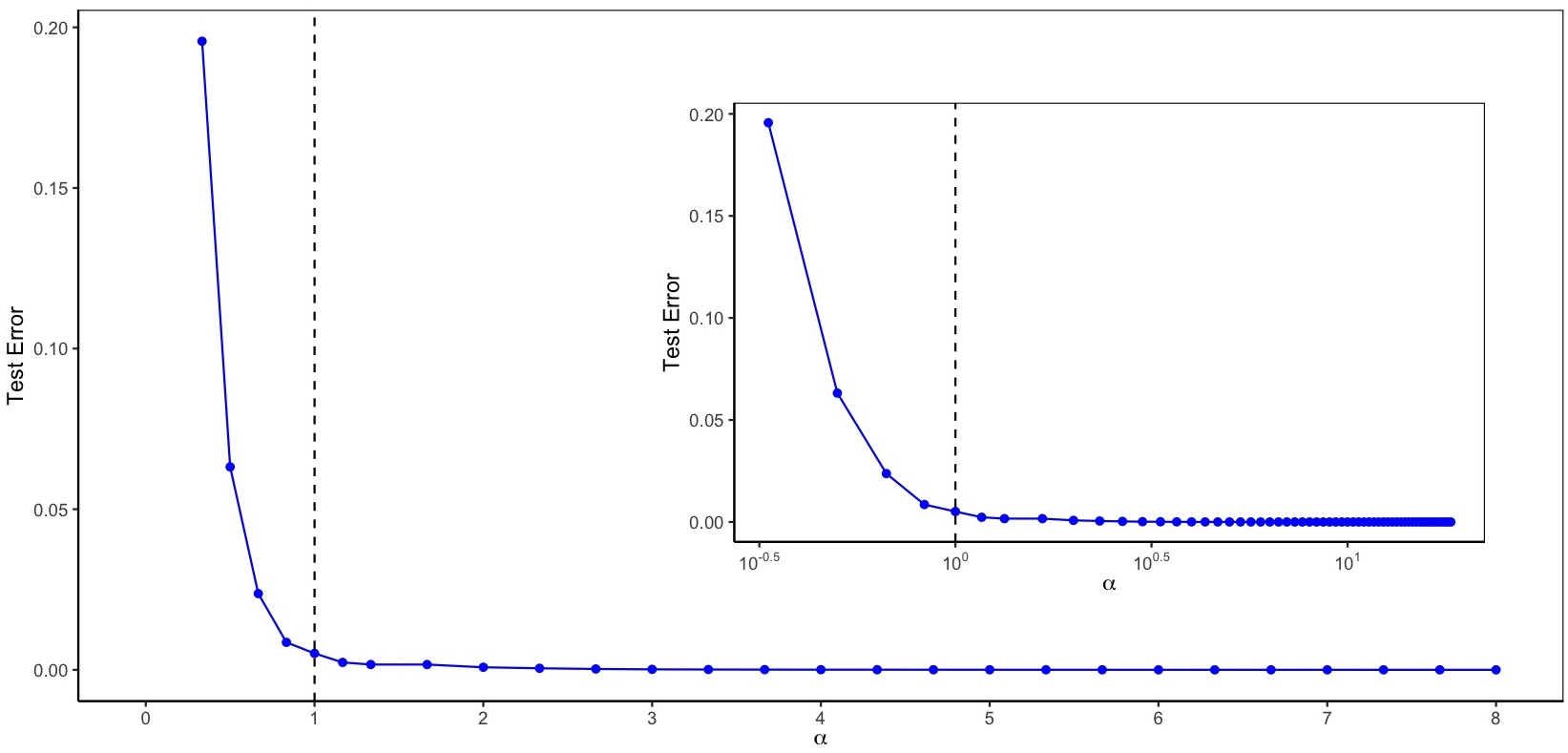}}
\end{figure}
Note that similar double descent behaviors using real datasets CIFAR-10 and MINST have been demonstrated for different types of models in the works \cite{DRBK} and \cite{LC}.\\

After observing the phenomenon of double descent in above binary classification model, next we aim to provide a theoretical foundation for this behavior by analyzing it within the context of a simple two-layer neural network architecture. This approach helps us understand how over-parameterization, under-fitting, and overfitting interact in neural networks, and offers insight into the mechanisms behind the observed empirical results.
\section{Two layer neural network with ReLU activation function}
\subsection{Problem setup}
We consider a two layer neural network for binary classification under supervised learning. For a given input vector $\textbf{x} \in \mathbb{R}^d$, the single output unit consists of a label $y\in \{-1,1\}$. The rectified linear unit or ReLU ($\sigma(z)=\max(0,z)$) is used as the activation function in the hidden layer.  This function is computationally simple and efficient and it is heavily used as the default activation function in deep neural networks. 

\subsection{Student model}

 The student model works as follows. The input layer loads the data into the neural network and it consists of $d$ nodes. Then the first layer with a single neuron computes a function of inputs from ${\mathbb{R}^d \text{ to }\mathbb{R}}$ as ${\textbf{x}_i^T\bm{\beta}}$ and sends it through the ReLU activation function. We use a  bias term $b$ to make the results more general.
 
\begin{equation}
\label{student}
f(\textbf{x}_i)=\sigma\left(\frac{\textbf{x}_i^T\bm{\beta}}{\sqrt{d}}+b\right).
\end{equation}
Here, $\bm{\beta}\in\mathbb{R}^d$ is the weight vector, $\textbf{x}_i\in\mathbb{R}^d$ is the $i^{th}$ feature vector for $i=1,..,n$, then $\textbf{x}_{i}^{T}\bm{\beta}=\sum_{j=1}^{d}x_{ij}^{T}\beta_j=x_{i1}\beta_{1}+...+x_{id}\beta_{d}, \in \mathbb{R}^d$. The $j^{th}$ feature for the $i^{th}$ training example is denoted as $x_{ij}$. The feature matrix denotes all the training cases along with their features as $\textbf{X}\in \mathbb{R}^{d\times n}$. Training the student model is done using a dataset with $n$ data points as $\mathcal{D}=\{(\textbf{x}_i,y_i) | \textbf{x}_i\in \mathbb{R}^d, y_i\in\{1,-1\},1\leq i\leq n\}$. \\

In the second layer which is the output layer consisting of a single node classifies the input $\textbf{x}_i$  into two labels as follows for all $i\in[n]$.

\begin{equation*}
\label{rule}
\hat y_i=
    \begin{cases}
        1 & \text{if } f(\textbf{x}_i)>0\\
        -1 & \text{if } f(\textbf{x}_i)\leq 0.
    \end{cases}
\end{equation*}

\subsection{Data generation - teacher model}

We study supervised binary classification under the following data distribution for feature vector $\textbf{x}_{i}\in \mathbb{R}^{d}$ using the class labels $y_{i}\in \{-1,1\}$. We incorporate two independent Gaussian vectors as $\bm{\eta}$ and $\bm{\epsilon}_i$ in $ \mathbb{R}^{d}$ which have components from $\mathcal{N}(0,1)$. In particular, for each $i \in [n]$, a given data point $\textbf{x}_i$ is generated as, 

\begin{equation}\label{teacher}
\textbf{x}_{i}= \frac{\bm{\eta}}{\sqrt{d}} y_{i} + \bm{\epsilon}_{i}.
\end{equation}

Each data point relates to one of the two class labels $\{-1,1\}$ with probabilities $\rho_{-1}$ and $\rho_{1}$ such that $\rho_{-1}+\rho_{1}=1$. We define two Gaussian clusters located at $\frac{\bm{\eta}}{\sqrt{d}}$ and $-\frac{\bm{\eta}}{\sqrt{d}}$. Since our data set consists of $n$ data points,  we expect to have $n\rho_{1}$ and $n\rho_{-1}$ points respectively in two clusters. Under this setting the feature vector $\textbf{x}_i\in \mathbb{R}^{d}$ is a Gaussian vector with mean $\textbf{0}$. 

\subsection{Loss and risk}

Our study is valid for margin based convex loss functions, such as square loss function. We evaluate the classification performance of the network $f$ by the \textbf{empirical risk} $\hat R_{n}(\bm{\beta})$ subject to a margin based loss function $l:\mathbb{R}\longrightarrow\mathbb{R}$ in binary classification. To overcome overfitting from too large weights for unseen data, we add $l_2$ regularization term to the empirical risk.

\begin{equation}
\label{loss}
\hat R_{n}(\bm{\beta})=\sum_{i=1}^{n}l\left(y_i\sigma\left (\frac{\textbf{x}_i^T\bm{\beta}}{\sqrt{d}}+b\right)\right)+\frac{\lambda}{2}||\bm{\beta}||_2^2.
\end{equation}

The non-negative tuning parameter $\lambda$ is used to control the balance between overfitting and under-fitting.  Most popular margin based loss functions are, square loss: $l(x)=\frac{1}{2}(1-x)^2$, hinge loss: $l(x)=\max\{0,1-x\}$ and logistic loss: $l(x)=\log(1+e^{-x})$. 

\subsection{Asymptotic setting and fixed quantities} 

We are interested in high dimensional setting where $n$ and $d$ $\longrightarrow \infty$ while preserving the ratio $\alpha=n/d$ fixed. Furthermore we will observe changes that are possible to happen with large sample sizes and higher dimensions as well as the ratios in between them. To achieve an accurate analytical formula for generalization error, we choose two non-negative quantities $r$ and $s$ having the following interpretations. We require each $\beta_j$ to be bounded for all $1\leq j\leq d$ and define,
\begin{equation}
\label{r and s}
r=\frac{1}{d}||\bm{\beta}||_2^2 \;\;\; \text{and} \;\;\; s=\frac{1}{d}\bm{\beta}^T\bm{\eta}.
\end{equation}
Later on we will restrict our attention to the domain $s^2\le r$ (see Section  \ref{train}). 

\section{Generalization}\label{t1}

We consider a new sample data $(\textbf{x}_N,y_N)$ and we quantify the test error (deterministic) of the model using $\bm{\hat\beta}=\bm{\beta}(\textbf{X},\textbf{y})$ where $(\textbf{X},\textbf{y})$ is our original training sample. Predicted value of the new data is decided using the classification rule $\hat{f}(\textbf{x}_N)=sgn(\sigma(\frac{\textbf{x}_N^T\bm{\hat\beta}}{\sqrt{d}}+b))$ where $sgn(z)=1  \:\: \text{if} \:\:  z>0$ and $sgn(z)=-1 \:\: \text{if}  \:\: z\leq0$ for any $z\in\mathbb{R}$. Let $\bm{\epsilon}_N$ have the components taken from $\mathcal{N}(0,1)$ pairing with the teacher model introduced in (\ref{teacher}). The generalization error is defined as the expectation of getting a misclassified output which is calculated using the indicator function.
\begin{equation*}
\begin{split}
&R(\bm{\hat\beta})=\mathbb{E}\left[ \mathds{1} (\hat f(\textbf{x}_N)\neq y_N)\right] 
\end{split}
\end{equation*}

\begin{theorem}\label{The1}
The test/generalization error of the two layer neural network model defined in (\ref{student}) under the fixed quantities $r$ and $s$ introduced in (\ref{r and s}) is given by,
\begin{equation}
\label{test error}
R(\bm{\hat\beta})=1-\rho_{1}\Phi\left(\frac{s+b}{\sqrt{r}}\right)-\rho_{-1}\Phi\left(\frac{s-b}{\sqrt{r}}\right),
\end{equation}
where $\Phi$ is the cumulative distribution function of standard normal distribution, $b$ is the bias term and $\rho_{-1}$ and $\rho_{1}$ relate to the probabilities of getting 1 or -1 in classification.
\end{theorem}
\begin{proof}
Observe that, $R(\bm{\hat\beta})=\mathbb{E}\left[ \mathds{1} (\hat f(\textbf{x}_N)\neq y_N)\right] 
=\mathbb{P}\left[\hat f(\textbf{x}_N)\neq y_N\right]$.  As $y_N$ takes the values $-1$ and $1$ with probabilities $\rho_{-1}$ and $\rho_1$, respectively,  we have\\
 \begin{align*}
&R(\bm{\hat\beta})=\mathbb{P}\left[\text{sgn}\Big(\sigma\Big(\frac{\textbf{x}_N^T\bm{\hat\beta}}{\sqrt{d}}+b\Big)\Big)\neq y_N\right]\\
&=\rho_1 \mathbb{P}\left[\text{sgn}\Big(\sigma\Big(\frac{\textbf{x}_N^T\bm{\hat\beta}}{\sqrt{d}}+b\Big)\Big) \neq  1 \right] +\rho_{-1}\mathbb{P}\left[\text{sgn}\Big(\sigma\Big(\frac{\textbf{x}_N^T\bm{\hat\beta}}{\sqrt{d}}+b\Big)\Big) \neq -1\right]\\
&= \rho_1 \mathbb{P}\left[\frac{\textbf{x}_N^T\bm{\hat\beta}}{\sqrt{d}}+b \leq 0 \right] +\rho_{-1}\mathbb{P}\left[\frac{\textbf{x}_N^T\bm{\hat\beta}}{\sqrt{d}}+b >0 \right].
\end{align*}
Next, use the teacher model (\ref{teacher}) to simplify the last line.  Also notice that $\bm\epsilon_{N}^T\bm\hat\beta$  follows Gaussian distribution with mean 0 and variance $rd$. Then using the definitions given in \ref{r and s}, we have the following.
\begin{equation*}
\begin{split}
&R(\bm{\hat\beta})= \rho_1 \mathbb{P}\left[\left(\frac{\bm{\eta}}{\sqrt{d}}+\bm{\epsilon_N}\right)^T\bm{\hat\beta}+b\sqrt{d} \leq 0 \right] +\rho_{-1}\mathbb{P}\left[\left(\frac{-\bm{\eta}}{\sqrt{d}}+\bm{\epsilon_N}\right)^T\bm{\hat\beta}+b\sqrt{d} >0 \right]\\
&=\rho_1 \mathbb{P}\left[\bm{\epsilon}_N^T\bm{\hat{\beta}} \leq -\sqrt{d}(s+b) \right]+ \rho_{-1} \mathbb{P}\left[\bm{\epsilon}_N^T\bm{\hat{\beta}} \geq \sqrt{d}(s-b) \right]\\
&=\rho_{1}\Phi\left(\frac{-(s+b)}{\sqrt{r}}\right)+\rho_{-1}\left(1-\Phi\left(\frac{s-b}{\sqrt{r}}\right)\right).
\end{split}
\end{equation*}
\end{proof}
We see that the generalization error depends on the values of $s$ and $r$ along with  the probabilities $\rho_{-1}$ and $\rho_{1}$. Moreover, when the bias term $b=0$, the generalization error depends on $s$ and $r$ only. In our model we assume that the ratio $\alpha=n/d$ is fixed as $n, d \to \infty$, and in Theorem \ref{mainT} we find the asymptotic values $r^*, s^*$ and $b^*$ of $r, s$ and $b$, respectively, as $n, d\to\infty$. 

\section{Regularized empirical risk}\label{train}
In this theoretical work, we do not follow any training algorithms as in decision trees, support vector machines, and logistic regression (\cite{BG}, \cite{M}, \cite{BG2}) or iterative optimization procedures like gradient descent and stochastic gradient descent (\cite{AKL}, \cite{HKL}). During the procedure, we feed the training data generated from the teacher model (\ref{teacher}) to the student model (\ref{student}). We expect to have minimal empirical loss and we solve the empirical risk minimization as an optimization problem.  We use Legendre transformation and the Convex Gaussian Min-max Theorem (CGMT) \cite{TOH2} to find a theoretical lower bound for the local training loss and avoid computing the exact local loss.\\


Our goal is to minimize the empirical risk in (\ref{loss}) to achieve an analytical formula for training loss subject to asymptotic settings in high dimension. First we define the concept of ``local training loss" $L_\lambda(r,s)$ using the fixed values $r, s$ and  regularization parameter $\lambda$:
\begin{equation}
\label{min problem}
\begin{split}
L_\lambda(r,s):= \min_{\bm{\beta}} \quad & \frac{1}{d}\sum_{i=1}^{n}l\left(y_i\sigma\left (\frac{\textbf{x}_i^T\bm{\beta}}{\sqrt{d}}+b\right)\right)+\frac{\lambda}{2d}||\bm{\beta}||_2^2,\\
\textrm{subject to} \quad & r=\frac{1}{d}||\bm{\beta}||_2^2,\\
&  s=\frac{1}{d}\bm{\beta}^T\bm{\eta}.
\end{split}
\end{equation}
It is easy to see that $l_2$ norm of standard Gaussian random vector is approximately equal to the square root of its dimension when the dimension is large enough. Hence by Cauchy-Schwartz inequality, we have 
 \begin{equation*}
|s|=\frac{|\bm{\beta}^T\bm{\eta}|}{d}\leq \frac{||\bm{\beta}||_2}{\sqrt{d}}\frac{||\bm{\eta}||_2}{\sqrt{d}}\approx\frac{\sqrt{rd}}{\sqrt{d}}\frac{\sqrt{d}}{\sqrt{d}}=\sqrt{r}.
\end{equation*}
Then $s^2\leq r$ holds in high dimensional settings and we define the ``global training loss" with the constraint $s^2\leq r$ as below.
\begin{equation}
\label{global training loss}
L_{\lambda}^*:=\min_{s^2\leq r} \quad  L_{\lambda}(r,s).
\end{equation}
Our next steps include finding a deterministic function for local training loss while keeping the ratio $\alpha=n/d$ fixed.
\subsection{Regularized empirical risk minimization procedure}

If we omit the constraints, the minimization problem in (\ref{min problem}) can be written as
\begin{equation}\label{ori}
\begin{split}
L_\lambda(r,s)&= \min_{\bm{\beta}} \;\; \Big\{ \frac{1}{d}\sum_{i=1}^{n}l\left(y_i\sigma\left (\frac{\textbf{x}_i^T\bm{\beta}}{\sqrt{d}}+b\right)\right)+\frac{\lambda}{2d}||\bm{\beta}||_2^2\Big\}\\
&=\min_{\bm{\beta}} \;\; \Big\{\frac{1}{d}\sum_{i=1}^{n}l\left(\frac{y_i}{2\sqrt{d}}\left(\textbf{x}_i^T\bm{\beta}+b\sqrt{d}+|\textbf{x}_i^T\bm{\beta}+b\sqrt{d}|\right)\right)+\frac{\lambda}{2d}||\bm{\beta}||_2^2\Big\}.
\end{split}
\end{equation}
In the second line above we have used the idea that the ReLU function can be represented as $\sigma(z)=(z+|z|)/2$. Solving this minimization problem is complicated and the derivative of the absolute value term does not exist when it is zero. Hence we use the Convex Gaussian Min-max Theorem (CGMT) (\cite{TOH2}, see also \cite{TOH}) to handle this minimization problem. For this purpose, we should rewrite the minimization problem on $\bm\beta$ as a combination of a maximization problem on a new variable ($\textbf{u}\in \mathbb{R}^n$) followed by the original minimization. This is a min-max problem since we deal with both minimization and maximization on two different variables. 

To convert the original problem (\ref{ori}) to a min-max problem, we utilize Legendre transformation for the convex loss function $l(\cdot)$ and rewrite it as a maximization problem (see supplementary
material 1 Section S.1 for more details).. 

Let $\tilde l(\cdot)$ be the Legendre transformation of the convex loss function $l(\cdot)$. Then we use Legendre transformation for each $i$, $1\leq i \leq n$ and rewrite $L_{\lambda}(r,s)$ as the following. 
\begin{equation*}\label{o2}
L_\lambda(r,s)= \min_{\bm{\beta}} \;\; \frac{1}{d}\sum_{i=1}^{n} \max_{u_i} \; \Big\{\frac{u_iy_i}{2\sqrt{d}}\left(\textbf{x}_i^T\bm{\beta}+b\sqrt{d}+|\textbf{x}_i^T\bm{\beta}+b\sqrt{d}|\right)-\tilde l(u_i)\Big\}+\frac{\lambda r}{2}.
\end{equation*}
The sum of the maximums with respect to each $u_i$ is the same as the maximum of the sum with respect to $\textbf{u}$. Together with the teacher model in (\ref{teacher}) and constraints in (\ref{r and s}), we have
 \begin{equation*}
L_\lambda(r,s)=
\frac{\lambda r}{2}+ \min_{\bm{\beta}} \; \max_{\textbf{u}}\; \frac{1}{d} \sum_{i=1}^{n} \left\{ \frac{u_is}{2}-\tilde l(u_i)+\frac{u_iy_ib}{2}+\frac{u_iy_i}{2\sqrt{d}}|\textbf{x}_i^T\bm{\beta}+b\sqrt{d}|+\frac{u_iy_i\bm{\epsilon}_i^T\bm{\beta}}{2\sqrt{d}}\right\}.
\end{equation*}
Next we use a standard normal vector $\bm\zeta_i^T$ to substitute the standard normal vector $y_i\bm\epsilon_i^T$.  
 \begin{equation}
\label{po}
L_\lambda(r,s)=\frac{\lambda r}{2}+ \min_{\bm{\beta}} \;\max_{\textbf{u}}\; \frac{1}{d} \sum_{i=1}^{n} \left\{ \frac{u_i(s+y_ib)}{2}-\tilde l(u_i)+\frac{u_iy_i}{2\sqrt{d}}|\textbf{x}_i^T\bm{\beta}+b\sqrt{d}|+\frac{u_i\bm{\zeta}_i^T\bm{\beta}}{2\sqrt{d}}\right\}.
\end{equation}

According to CGMT, we shall call (\ref{po}) as the primary optimization problem (PO) derived from the original minimization problem in (\ref{ori}). Next, considering the convexity of $\tilde l(u_i)$ and the $d$ dimensional Gaussian random vector $\bm\zeta$, we define the auxiliary optimization problem (AO) which is denoted by $\tilde L_\lambda(r,s)$. For $\bm{\beta}\in\mathbb{R}^d$ and $\textbf{u}\in\mathbb{R}^n$, let $\textbf{g}\sim \mathcal{N}(0,I_n)$ and $\textbf{h}\sim \mathcal{N}(0,I_d)$ be two Gaussian vectors in $\mathbb{R}^n$ and $\mathbb{R}^d$ respectively. Then $\tilde L_\lambda(r,s)$ is defined as 
\begin{equation}
\label{ao1}
\tilde L_\lambda(r,s)=\frac{\lambda r}{2}+ \min_{\bm{\beta}} \;\;\max_{\textbf{u}} \Big\{\psi(\bm{\beta},\textbf{u}) +\frac{\sqrt{r}}{2d}\sum_{i=1}^nu_ig_i+\frac{||\textbf{u}||_2\textbf{h}^T\bm{\beta}}{2d\sqrt{d}}\Big\},
\end{equation}
where
\begin{equation}\label{si}
\psi(\bm{\beta},\textbf{u}):=\frac{1}{d} \sum_{i=1}^{n} \left(\frac{u_{i}(s+y_ib)}{2}-\tilde l(u_i)+\frac{u_iy_i}{2\sqrt{d}}|\textbf{x}_i^T\bm{\beta}+b\sqrt{d}|\right).
\end{equation}
According to the CGMT theorem, for any constant $c\in \mathbb{R}$, we have 
\begin{equation}
\label{cgmt 1}
\mathbb{P}(L_\lambda(r,s)<c) \leq 2\mathbb{P} ( \tilde L_\lambda(r,s)<c ).
\end{equation}
\begin{flalign*}
\text{Then, }\tilde L_\lambda(r,s)\geq\frac{\lambda r}{2}+ \min_{\bm{\beta}} \;\;\max_{\textbf{u}, u_iy_i>0} \Big\{\psi(\bm{\beta},\textbf{u}) +\frac{\sqrt{r}}{2d}\sum_{i=1}^nu_ig_i+\frac{||\textbf{u}||_2\textbf{h}^T\bm{\beta}}{2d\sqrt{d}}\Big\}.&&
\end{flalign*}

Using the convexity of the loss function and following the standard application procedures (see \cite{MF} and page 5 of supplementary reading \cite{MF}) of the CGMT theorem we get, 

 \begin{equation}
\label{cgmt 2}
\mathbb{P} ( \tilde L_\lambda(r,s)<c ) \leq \mathbb{P} ( \omega_{\lambda}^{(d)}(r,s)<c ).
 \end{equation}
Now, combining the inequalities in (\ref{cgmt 1}) and (\ref{cgmt 2}) yields 
 \begin{equation}
 \label{cgmt 3}
 \mathbb{P}(L_\lambda(r,s)<c) \leq 2\mathbb{P} ( \omega_{\lambda}^{(d)}(r,s)<c ).
 \end{equation}

 Next we solve the maximization problem on Legendre transformation variable $\textbf{u}\in \mathbb{R}^n$ in  $\omega_{\lambda}^{(d)}(r,s)$. We use the Lagrange multiplier method for this maximization as it is a simple optimization problem with one constraint. Let $p^*$ to be the optimal solution of the maximization problem
\begin{equation*}
\label{max u}
p^*:=\max_{\substack{\textbf{u}, \\ u_iy_i>0}} \left\{\frac{1}{d}\sum_{i=1}^{n} \Big(\frac{u_i}{2}(s+y_ib+\sqrt{r}g_i)-\tilde l(u_i)\Big) +\Big(\frac{s\bm{\eta}^T\textbf{h}}{d}-\sqrt{r-s^2}\Big)\frac{||\textbf{u}||_2}{2\sqrt{d}}\right\}.
\end{equation*}
Let $\tau_i$ be the non-negative Lagrange multipliers associated with $u_iy_i>0$ for $1\le i \le n$. Then the Lagrangian function can be written as
 \begin{align*}
 \label{lagrangian 1}
L(\textbf{u},\bm{\tau})= \frac{1}{d}\sum_{i=1}^{n} \Big(\frac{u_i}{2}(s+y_ib+\sqrt{r}g_i)-\tilde l(u_i)\Big)+\Big(\frac{s\bm{\eta}^T\textbf{h}}{d}-\sqrt{r-s^2}\Big)\frac{||\textbf{u}||_2}{2\sqrt{d}} +\frac{1}{d}\sum_{i=1}^n\tau_iu_iy_i.
\end{align*}
Then, $ p^*= \max_{\substack{\tau_i\geq 0, \textbf{u}}} \; L(\textbf{u},\bm{\tau})$. Let the optimal values of $(\textbf{u},\bm{\tau})$ be $(\textbf{u}^*,\bm{\tau}^*)= \arg\max_{\substack{\tau_i\geq 0, \textbf{u}}} \; L(\textbf{u},\bm{\tau}) $. Then following Theorem 18.5 in \cite{SB} with above $L(\textbf{u},\bm{\tau})$, we have 
\begin{enumerate}
\item[(a)] $\dfrac{\partial L(\textbf{u}^*,\bm\tau^*)}{\partial u_i^*}=\frac{1}{d}\Big(\frac{1}{2}(s+y_ib+\sqrt{r}g_i)-\tilde l^{\prime}(u^*_{i})\Big)+\Big(\frac{s\bm{\eta}^T\textbf{h}}{d}-\sqrt{r-s^2}\Big)\frac{\textbf{u}^*_i}{2\sqrt{d}||\textbf{u}^*||_2}+\frac{\tau^*_{i}}{d}y_i=0,$
\item[(b)]  $ \tau_{i}^*(u_{i}^*y_i)=0, $  \;\;\;\ (c)  $ \tau_{i}^*\geq 0$ \;\;\;\; (d) $ u_{i}^*y_i>0, \quad \text{for all} \quad 1 \leq i \leq n. $
\end{enumerate} 
From (b) and (d) we have, $\tau_{i}^*=0$ for $1\leq i\leq n$ and this agrees with (c). Along with this, we rewrite (a) as,
\begin{equation}
\label{max u sol}
\Big(\sqrt{r-s^2}-\frac{s\bm{\eta}^T\textbf{h}}{d}\Big)\frac{\sqrt{d}}{2||\textbf{u}^*||_2} \;u^*_i+\tilde l^{\prime}(u^*_{i})=\frac{1}{2}(s+y_ib+\sqrt{r}g_i).
\end{equation}
This creates a system of equations on $u_i^*, i=1, \cdots, n$. Recall that $\textbf{g}=(g_1,\cdots,g_n)$ is a random vector with i.i.d. standard normal entries and $y_i=\pm 1$ is independent of $g_i$. Then we have the solution to the maximization problem.
\begin{equation*}
\label{p star}
p^*=\frac{1}{d}\sum_{i=1}^{n} \Big(\frac{u_i^*}{2}(s+y_ib+\sqrt{r}g_i)-\tilde l(u_i^*)\Big) +\Big(\frac{s\bm{\eta}^T\textbf{h}}{d}-\sqrt{r-s^2}\Big)\frac{||\textbf{u}^*||_2}{2\sqrt{d}}.
\end{equation*}
We plug $p^*$ back into $\omega_{\lambda}^{(d)}(r,s) $ and simplify it using the relationship in (\ref{max u sol}).
\begin{align}
\omega_{\lambda}^{(d)}(r,s) 
&= \frac{\lambda r}{2}+\frac{1}{d}\sum_{i=1}^{n} (u_i^*\tilde l^{\prime}(u_i^*) -\tilde l(u_i^*)) \label{omega solved}
\end{align}
This $\omega_{\lambda}^{(d)}(r,s)$ in (\ref{omega solved}) is a lower bound for the AO problem. This quantity depends on $\textbf{u}^*$ which satisfies (\ref{max u sol}). \\

\noindent\underline{\textbf{Scalar change of variables by the substitution $v_i=\tilde l^{\prime}(u_i^*)$}}\\


Here we introduce a scalar change of variables by $v_i=\tilde l^{\prime}(u_i^*)$. Properties of Legendre transformation help to verify the following two identities and the detailed steps are shown in supplementary material 1 Section S.2.
\begin{equation}
\label{scalar change}
\begin{split}
u_i^*= l^{\prime}(v_i)\\
 u_i^* \tilde l^{\prime}(u_i^*)-\tilde l(u_i^*)=l(v_i).
\end{split}
\end{equation}
For each $i\in[n]$, $v_i\in\mathbb{R}$ and we denote them by vector $\textbf{v}\in\mathbb{R}^n$. Moreover by $l^{\prime}(\textbf{v})$, we mean the element-wise derivative on the vector $\textbf{v}$. The new vector $\textbf{v}$ changes the expression in (\ref{max u sol}) to the following.
\begin{equation}
\label{u to v}
\Big(\sqrt{r-s^2}-\frac{s\bm{\eta}^T\textbf{h}}{d}\Big)\frac{\sqrt{d}}{2||l^{\prime}(\textbf{v})||_2} \;l^{\prime}(v_i)+v_{i}=\frac{1}{2}(s+y_ib+\sqrt{r}g_i).
\end{equation}
\begin{flalign}
\label{gamma}
\text{For easy reference we denote, }\gamma=\Big(\sqrt{r-s^2}-\frac{s\bm{\eta}^T\textbf{h}}{d}\Big)\frac{\sqrt{d}}{2||l^{\prime}(\textbf{v})||_2}.&&
\end{flalign}

\begin{flalign}
\label{rel 1}
\text{Hence (\ref{u to v}) can be re-written as, }\gamma l^{\prime}(v_i)+v_{i}=\frac{1}{2}(s+y_ib+\sqrt{r}g_i).&&
\end{flalign}
Applying the second relation in (\ref{scalar change}) to $\omega_{\lambda}^{d}(r,s)$ in (\ref{omega solved}) we get the following expression for $\omega_{\lambda}^{d}(r,s)$.
  \begin{equation}
  \label{omega v d}
 \begin{split}
\omega_{\lambda}^{(d)}(r,s)&=\frac{\lambda r}{2}+\frac{\alpha}{n}\sum_{i=1}^{n} l(v_i).
 \end{split}
\end{equation}
Now the calculations only require a margin based loss function and future calculations does not require finding the convex conjugates of the loss functions. Using the CGMT, we can show that $\omega^{(d)}_{\lambda}(r,s)$ can be used as a candidate to observe the asymptotic behavior of the global training loss (see supplementary material 1 Section S.3 for more details). Since we already have an expression for $\omega^{(d)}_{\lambda}(r,s)$ in (\ref{omega v d}), first we minimize it to find  $\omega^*_{\lambda}(r,s)$  and finally consider the high dimensional behavior by sending $n, d \to \infty$.\\

\noindent\underline{\textbf{Minimizing $\omega^{(d)}_{\lambda}(r,s)$ to find $r, s \text{ and } b$}}\\

As explained in Theorem \ref{The1}, the generalization error depends on the fixed values of $r, s \text{ and } b$. Through the training procedure of the model, we find the fixed quantities  $r, s \text{ and } b$ that correspond to the minimum possible training error of the model. Hence we work on the minimization of $\omega_{\lambda}^{(d)}(r,s)$ in (\ref{omega v d}) with constraint $s^2\leq r$ to find $\omega^*_{\lambda}(r,s)$, which can be done by setting the derivatives of $\omega^{(d)}_{\lambda}(r,s)$ with respect to $r, s$ and $b$ to zero. We use the relationship introduced in (\ref{rel 1}) and the following relationship derived from (\ref{gamma}) to find the partial derivatives of $v$ and $\gamma$ as needed. 
\begin{equation}
\label{gamma^2}
4\alpha\gamma^2||l^{\prime}(\textbf{v})||_2^2=n\Big(\sqrt{r-s^2}-\frac{s\bm{\eta}^T\textbf{h}}{d}\Big)^2.
\end{equation}
More details on calculating the following derivatives are explained in the supplementary material
1 Section S.5.. The derivative of (\ref{omega v d}) with respect to $r$ is given by,
\begin{equation}\label{r final}
\frac{\alpha}{\sqrt{r}}\frac{1}{n}\sum_{i=1}^n g_i(w_i-2v_i)=-4\lambda \gamma+1-\frac{s\bm\eta^T\textbf{h}}{d \sqrt{r-s^2}},
\end{equation}
where $w_i=(s+y_ib+\sqrt{r}g_i)$ and $w_i$ follows a normal distribution with mean $s+y_ib$ and standard deviation $\sqrt{r}$ conditioned on $y_i$ for each $1\leq i\leq n$. The derivative of (\ref{omega v d}) with respect to $s$ is given by,
\begin{equation}\label{s final}
-\alpha\frac{1}{n}\sum_{i=1}^n(w_i-2v_i)=s+G,
\end{equation}
where $G=\sqrt{r-s^2}\frac{\bm\eta^T\textbf{h}}{d}-\frac{s^2\bm\eta^T\textbf{h}}{d\sqrt{r-s^2}}-s\left(\frac{\bm\eta^T\textbf{h}}{d}\right)^2 $. The derivative of (\ref{omega v d}) with respect to $b$ is given by,
\begin{equation}
\label{fixed b}
\sum_{i=1}^ny_i(w_i-2v_i)=0.
\end{equation}
For fixed $n$ and $d$, solving the equations (\ref{r final}), (\ref{s final}) and (\ref{fixed b}) may give us the optimal $r, s$ and $b$ values which satisfy $\omega^{*}_{\lambda}(r,s)$. However, these three equations cannot be solved in general. Hence we select a margin-based convex loss function and solve the equations. Since $\gamma$ is involved here, we use the relationship in (\ref{rel 1}) as $2v_i+2\gamma l^{\prime}(v_i)=w_i$ when needed.


Since we are interested in high dimensional behavior of $r, s$, and $b$, we look at the limiting behavior of the three equations when $n, d \to \infty$. Let $r^*, s^*, b^*$ and $\gamma^*$ be the values achieved by $r, s, b$ and $\gamma$ when $n, d \to \infty$.\\

\noindent\underline{\textbf{Finding the asymptotics of $r, s, b$ and $\gamma$}}\\

We consider the limits  of both sides of (\ref{r final}).
\begin{equation*}
\lim_{n,d\to \infty} \left( \frac{\alpha}{\sqrt{r}}\frac{1}{n}\sum_{i=1}^n g_i(w_i-2v_i)\right)=\lim_{n,d\to \infty} \left(-4\lambda \gamma+1-\frac{s\bm\eta^T\textbf{h}}{d \sqrt{r-s^2}}\right).
\end{equation*}
After simplifying both sides it shows the following asymptotic relationship.  Notice that the last term in right side goes to zero since $\frac{\bm\eta^T\textbf{h}}{d}\to 0$ by law of large numbers  when $d\to\infty$.
\begin{equation}
\label{r limit}
\frac{\alpha}{\sqrt{r^*}}\lim_{n,d\to\infty} \left(\frac{1}{n}\sum_{i=1}^n g_i(w_i-2v_i)\right)=-4\lambda \gamma^*+1.
\end{equation}
Next, we consider the limits of both sides  of (\ref{s final}). 
\begin{equation}
\label{s limit}
-\alpha \lim_{n,d\to\infty} \left(\frac{1}{n}\sum_{i=1}^n(w_i-2v_i)\right)=s^*.
\end{equation}
Notice that when $n,d\to \infty$, the term $G=\sqrt{r-s^2}\frac{\bm\eta^T\textbf{h}}{d}-\frac{s^2\bm\eta^T\textbf{h}}{d\sqrt{r-s^2}}-s\left(\frac{\bm\eta^T\textbf{h}}{d}\right)^2 $ also goes to zero since $\frac{\bm\eta^T\textbf{h}}{d}\to 0$ by law of large numbers. By  (\ref{fixed b}), we have 
\begin{equation}
\label{b limit}
\lim_{n,d\to\infty} \left(\frac{1}{n}\sum_{i=1}^ny_i(w_i-2v_i)\right)=0.
\end{equation}
Since $\gamma$ is involved in above expressions as function of $\textbf{v}$, we consider the high dimensional behavior of (\ref{gamma^2}). First we rewrite it using the expression (S14).
\begin{equation*}
\alpha \frac{1}{n}\sum_{i=1}^n(w_i-2v_i)^2=\left (\sqrt{r-s^2}-\frac{s\bm\eta^T\textbf{h}}{d}\right)^2.
\end{equation*} Then the limits give
\begin{equation}
\label{gamma limit}
\alpha\lim_{n,d\to\infty} \left( \frac{1}{n}\sum_{i=1}^n(w_i-2v_i)^2\right)=r^*-(s^*)^2.
\end{equation}
For a specific loss function, we can solve the above four equations (\ref{r limit}), (\ref{s limit}),(\ref{b limit}) and (\ref{gamma limit}) to find $r^*, s^*, \gamma^*$ and $b^*$. In summary, we have the following main result of the paper.
\begin{theorem}\label{mainT}
Let $f(\textbf{x}_i)=\sigma(\frac{\textbf{x}_i^T\bm{\beta}}{\sqrt{d}}+b)$ be a two-layer neural network with ReLU activation function. We minimize the empirical risk (\ref{loss}). Denote $r=\frac{1}{d}||\bm{\beta}||_2^2$ and $s=\frac{1}{d}\bm{\beta}^T\bm{\eta}$. Assume that data are generated from the teacher model in (\ref{teacher}),  the values of $\alpha$ and regularization parameter $\lambda$ are known, and $l$ is the square loss function. Let $w_i\sim\mathcal{N}(s+y_ib,r)$ conditional on $y_i=\pm 1$ and $g_i\sim\mathcal{N}(0,1)$ for all $i=1,\cdots, n$. $v_i\in \mathbb{R}$ satisfies the following relationship for any $\gamma>0$.
 \begin{equation}
 \label{t1 v}
\gamma l^{\prime}(v_i)+v_i=\frac{1}{2}w_i.
\end{equation}
 For  fixed ratio $\alpha$, the quantities $r, s, b$ and $
\gamma$ converge to fixed quantities $r^*, s^*, b^*$ and $ \gamma^*$ as $n, d\rightarrow \infty$. These fixed quantities are given by the solutions of the following equations.
\begin{equation}
\label{t1 r}
\frac{\alpha}{\sqrt{r^*}}\lim_{n,d\to\infty} \left(\frac{1}{n}\sum_{i=1}^n g_i(w_i-2v_i)\right)=-4\lambda \gamma^*+1,
\end{equation}
\begin{equation}
\label{t1 s}
-\alpha \lim_{n,d\to\infty} \left(\frac{1}{n}\sum_{i=1}^n(w_i-2v_i)\right)=s^*,
\end{equation}
\begin{equation}
\label{t1 b}
\lim_{n,d\to\infty} \left(\frac{1}{n}\sum_{i=1}^ny_i(w_i-2v_i)\right)=0,
\end{equation}
\begin{equation}
\label{t1 gamma}
\alpha\lim_{n,d\to\infty} \left( \frac{1}{n}\sum_{i=1}^n(w_i-2v_i)^2\right)=r^*-(s^*)^2.
\end{equation}
These quantities $r^*, s^*$ and $b^*$ are used to calculate the limit of generalization error of $f(\textbf{x}_i)$ in Theorem \ref{The1}, for a fixed $\alpha$ when $n, d\rightarrow \infty$.
\end{theorem}

\section{Theoretical generalization error curves for square loss }\label{curve sq}
Here we apply Theorem \ref{The1} and Theorem \ref{mainT} with square loss and analyze the theoretical generalization error as a function of $\alpha$ for different values of $\lambda$ and $\rho_1$. The calculation and the R code for the below work is presented in the supplementary material 1 Section S.6 and supplementary material 2 Section S.1 respectively. The regularization is used to balance over-fitting and under-fitting in the model. Our strategy prevents us from having $\lambda=0$ hence we use negligible $\lambda$ values to mean no regularization and vice-versa. Also, observe that the number of parameters in the model is same as the model dimension ($p=d$).
\begin{figure}[H]
\floatbox[{\capbeside\thisfloatsetup{capbesideposition={right,center},capbesidewidth=4cm}}]{figure}[\FBwidth]
{\caption{\small{Test error of model (\ref{student}) with square loss as a function of $\alpha$ with low regularization ($\lambda=10^{-5}$) and with $\rho_1=0.5$.}\label{fig1}}}
{\includegraphics[scale=0.4]{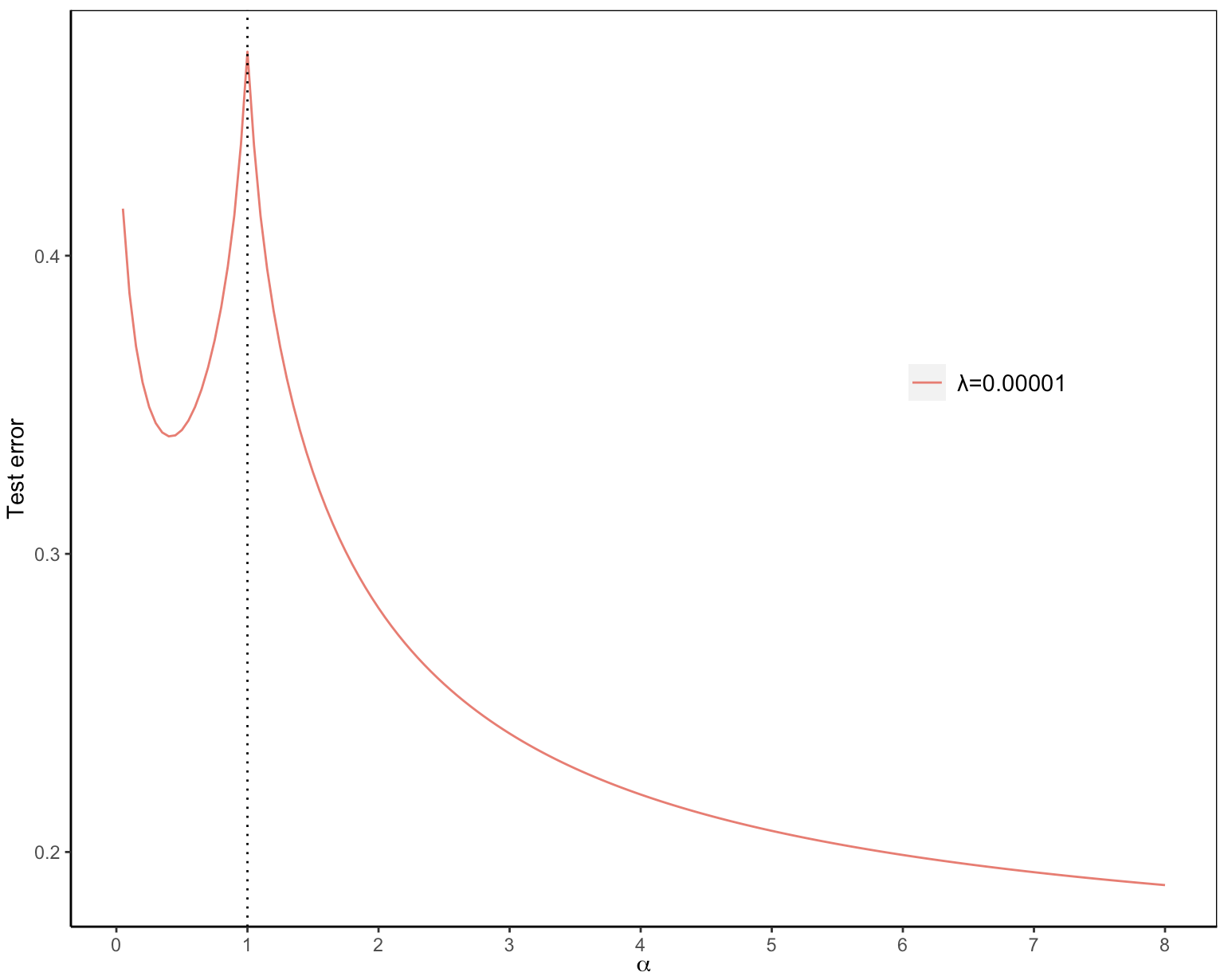}}
\end{figure}

Figure \ref{fig1} shows the test error as a function of $\alpha=n/d$ in the model (\ref{student}) under the square loss function. Here $\rho_1=\rho_{-1}=0.5$. Thus, the two classification groups have the same probability and we take the regularization parameter to be very small: $\lambda=10^{-5}$. As we can see, the test error first decreases, then increases again and reaches its peak when $\alpha=1$,  and later when $\alpha>1$, the test error steadily keeps decreasing.

\begin{itemize}
\item{When $\alpha<1$, i.e., when $n<d$, we have the over-parameterized region of the model in (\ref{student}). The test risk decreases first and then rises up again tracing a U-shaped curve. The first local minimum occurs before $\alpha=1$ and we identify this minimum as the sweet-spot.}
\item{Next, the increasing test error reaches its peak when $\alpha=1$, that is, when $n=d$. Hence, maximum test error occurs when $n=d$ and we identify this point as the interpolation threshold. }
\item{Finally, when $\alpha>1$, i.e., when $n>d$, the model enters the under-parameterized region and the test error decreases monotonically for increasing $\alpha$ values. With more data the model overfits resulting in a lower test error and the best performance of the model is achieved in this region agreeing with the classic idea ``more data is always better".}
\end{itemize}

So it is clear that we can jump from over-parameterized region to under-parameterized region by increasing $\alpha$ while observing the peak in between. The curve manifests the classical U-shaped curve in the over-parametrized region and the long plateau in the under-parametrized region. This is a noticeable difference between the double descent behavior observed as a function of model capacity ($n \text{ and } d$ are  fixed,  numbers of parameters are varying) and this ratio-based double descent behavior (the test error is a function of $\alpha=n/d$, with $n$ and $d$ going to infinity).

The over-parameterized region has its own local minima that corresponds to a better-performing model, and in under-parameterized region we have the flexibility to pick the best model since the test error is decreasing monotonically. For this specific binary classification model, the best model comes from the under-parameterized region and the test error of that model is comparatively smaller than any model coming from the over-parameterized region. We do not work on training of the model using iterative procedures like gradient descent and, therefore, we cannot comment on the position of global minima, as it depends on the composition of training data.

\subsection{Effect of regularization on double descent behavior}\label{regdd}
\begin{figure}[H]
\floatbox[{\capbeside\thisfloatsetup{capbesideposition={right,center},capbesidewidth=4cm}}]{figure}[\FBwidth]
{\caption{Test error of model (\ref{student}) for square loss as a function of $\alpha$ with $\rho_1=0.5$ and with varying regularization.}\label{fig2}}
{\includegraphics[scale=0.4]{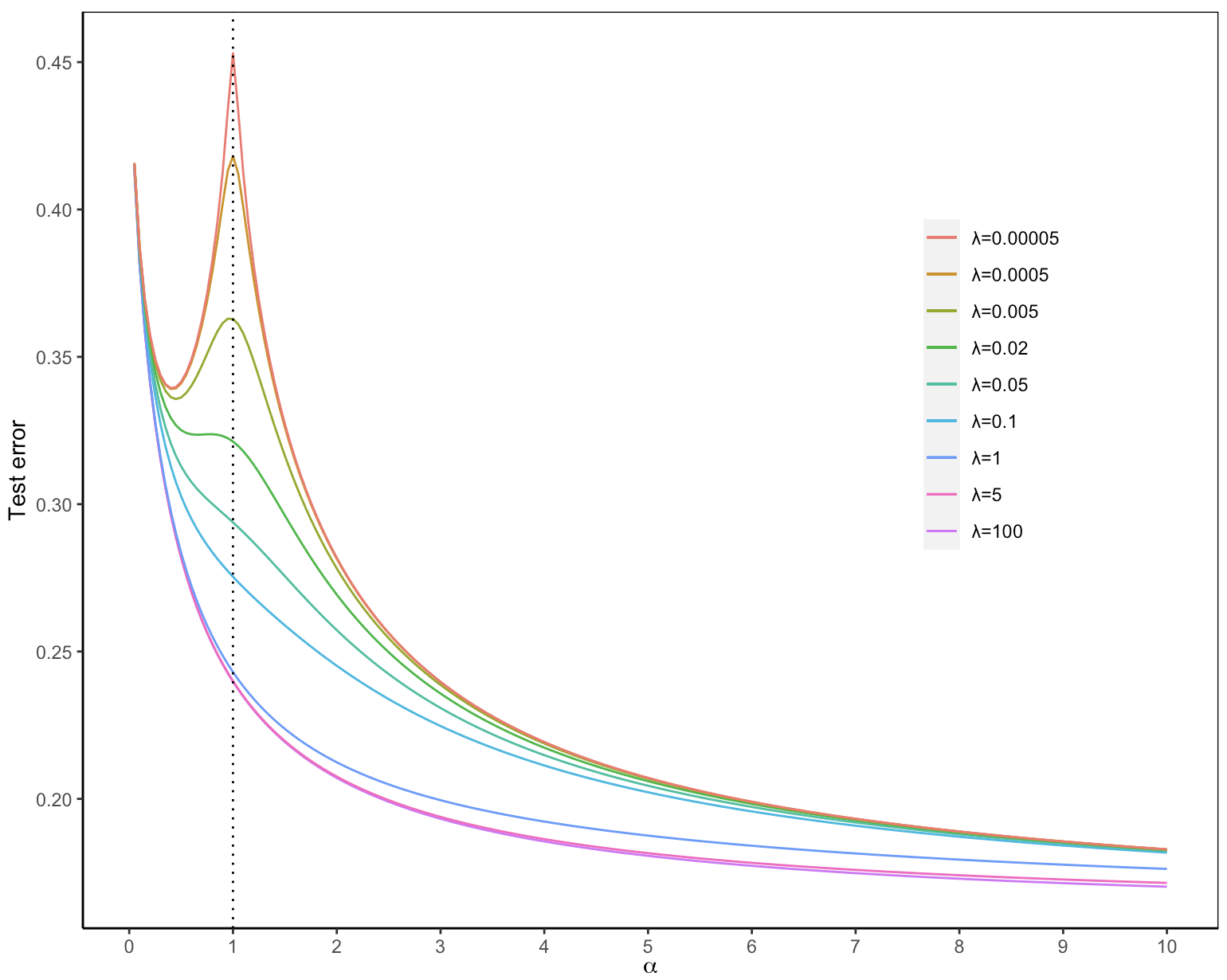}}
\end{figure}

Figure \ref{fig2} illustrates how the increasing regularization can smooth out the peak in the generalization error curve for the case $\rho_1=\rho_{-1}=0.5$. We did not allow $\lambda$ to be exactly zero for the numerical stability of the algorithm we followed. Peak is clearly visible with lower regularization like $\lambda=0.005$ and with further increase in regularization, the peak gets smoothed out and test error decreases monotonically. 

Moreover, this depicts the significance of having regularization in classification tasks to achieve better results. Similar studies done in \cite{NVKM} confirm that most under-regularized linear regression models observe this type of double descent curve. In this setting, when $\rho_1=0.5$, it seems that the higher $\lambda$ values, the lower test errors. However, it does not improve much if $\lambda$ is too large. We suggest to use $\lambda=5$ with equal class sizes. 


\subsection{Different $\rho_1$ values with weak and strong regularization}
 
Test error curves from the previous part correspond to equal cluster sizes with $\rho_1=0.5$. Now we study the double descent behavior for uneven cluster sizes. 

As illustrated in Figure \ref{fig3} below, when $\rho_1=0.7$, then, for lower regularization values, the test error starts at $0.3$ and then goes down monotonically, and the double descent behavior can be clearly observed. But when $\lambda\geq 1$, the test error keeps unchanged until a specific $\alpha$ value is reached and then starts decreasing monotonically as $\alpha$ gets bigger and bigger. Even though the higher regularization values acted similar on cases with $\rho_1=0.5$, when the cluster sizes are uneven, too much regularization does not yield favorable results.

\begin{figure}[H]
\floatbox[{\capbeside\thisfloatsetup{capbesideposition={right,center},capbesidewidth=4cm}}]{figure}[\FBwidth]
{\caption{Test error of model (\ref{student}) for square loss as a function of $\alpha$ with $\rho_1=0.7,$ and with varying regularization values.}\label{fig3}}
{\includegraphics[scale=0.37]{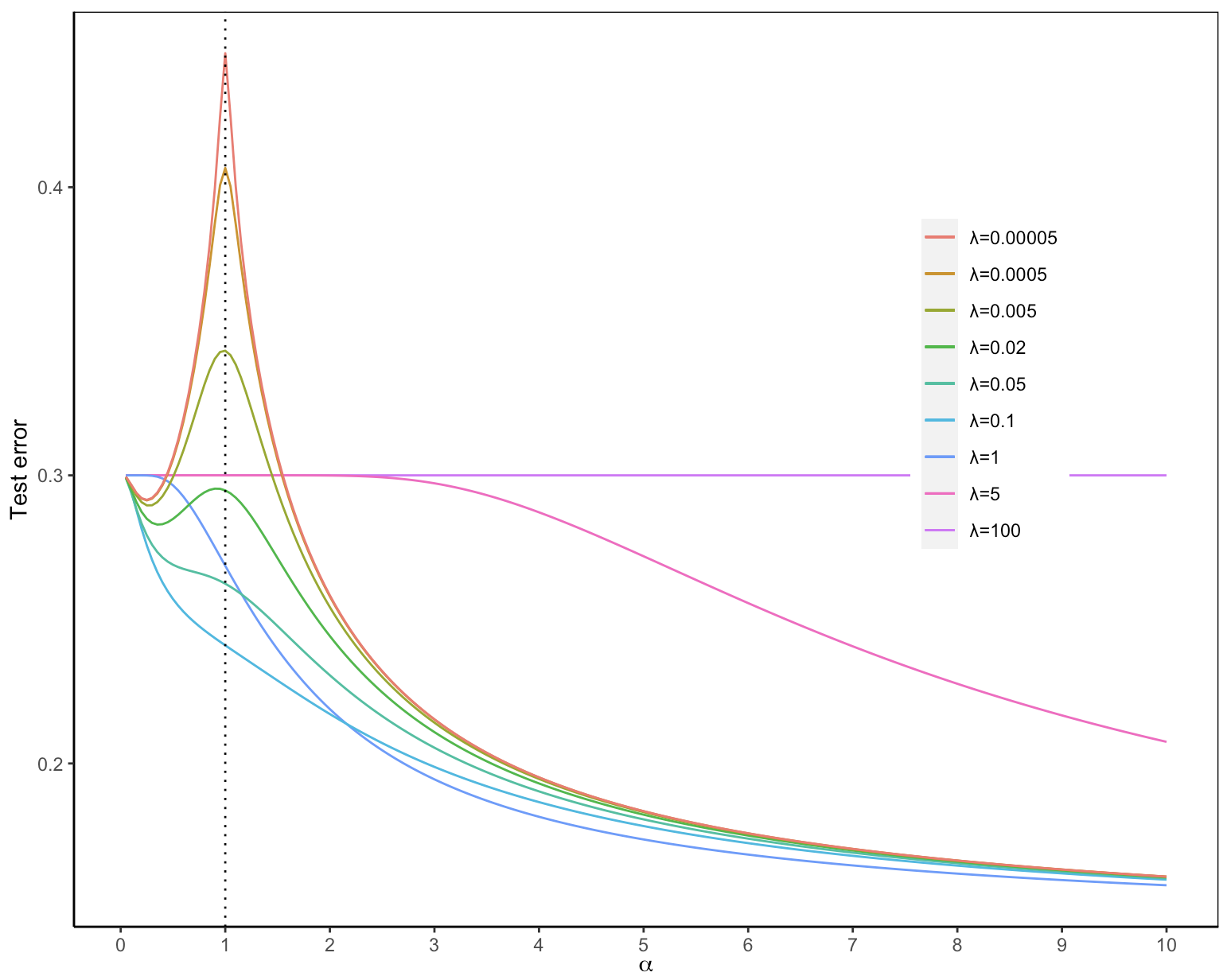}}
\end{figure}

\begin{figure}[H]
\floatbox[{\capbeside\thisfloatsetup{capbesideposition={right,center},capbesidewidth=4cm}}]{figure}[\FBwidth]
{\caption{Test error of model (\ref{student}) for square loss as a function of higher $\alpha$ values with $\rho_1=0.7$ and with varying regularization values.}\label{fig4}}
{\includegraphics[scale=0.4]{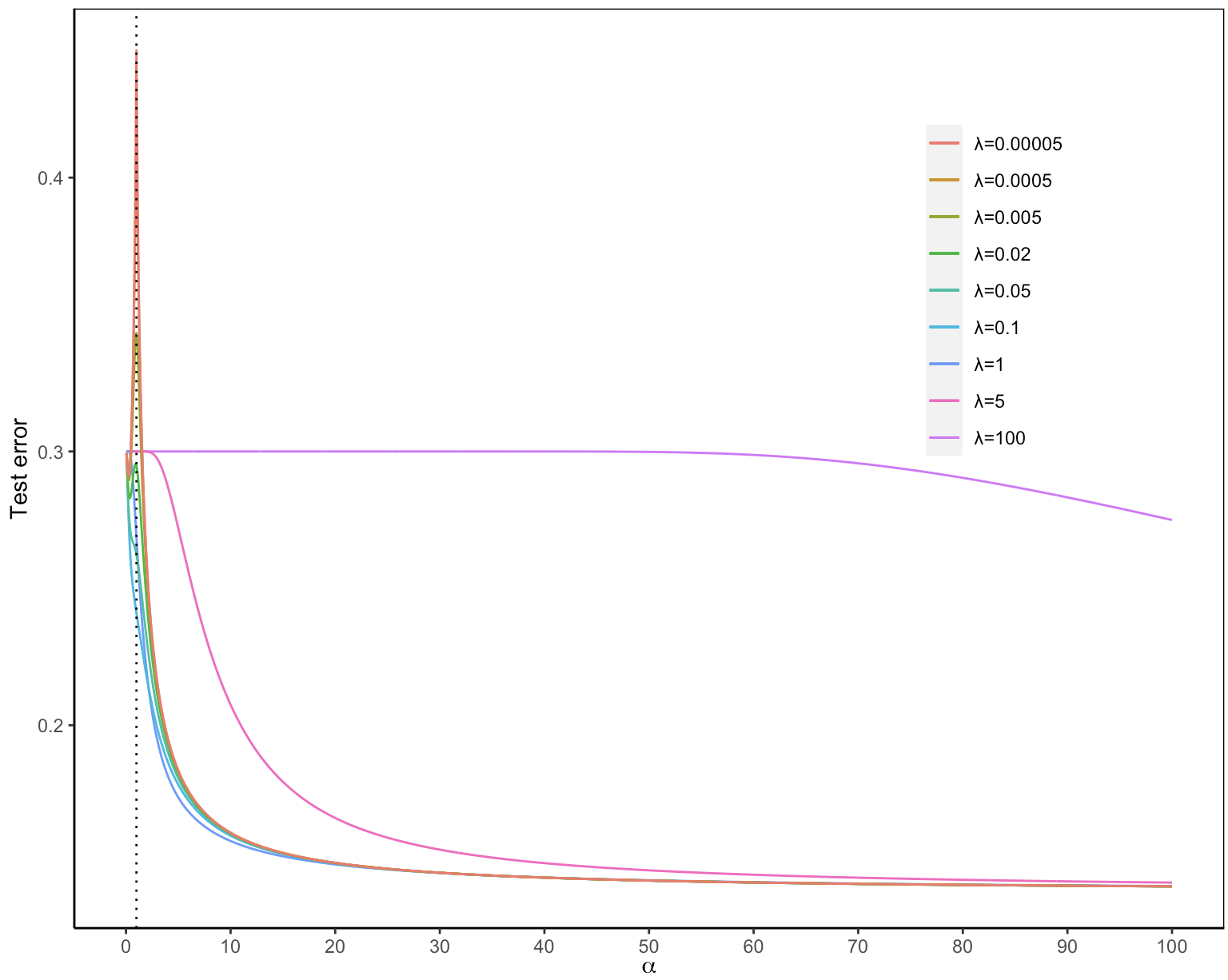}}
\end{figure}

Figure \ref{fig4} shows the same curves in Figure \ref{fig3} for higher $\alpha$ values to observe the downward trend for highly regularized models. We can see that even when $\lambda=100$, the test error starts decreasing when $\alpha\approx 60$. It is clear that the higher the regularization is, the higher values of $\alpha$ are needed to achieve a considerably lower test error, as the rate of decrease is very low. In summary, we shall have some suitable regularization ($0.1\le \lambda\le 1$) to have best performance.

\subsection{Test error of different cluster sizes for increasing regularization when $\alpha$ is fixed}

\begin{figure}[H]
    \centering
    \begin{minipage}{0.5\textwidth}
        \centering
        \includegraphics[width=1.\textwidth]{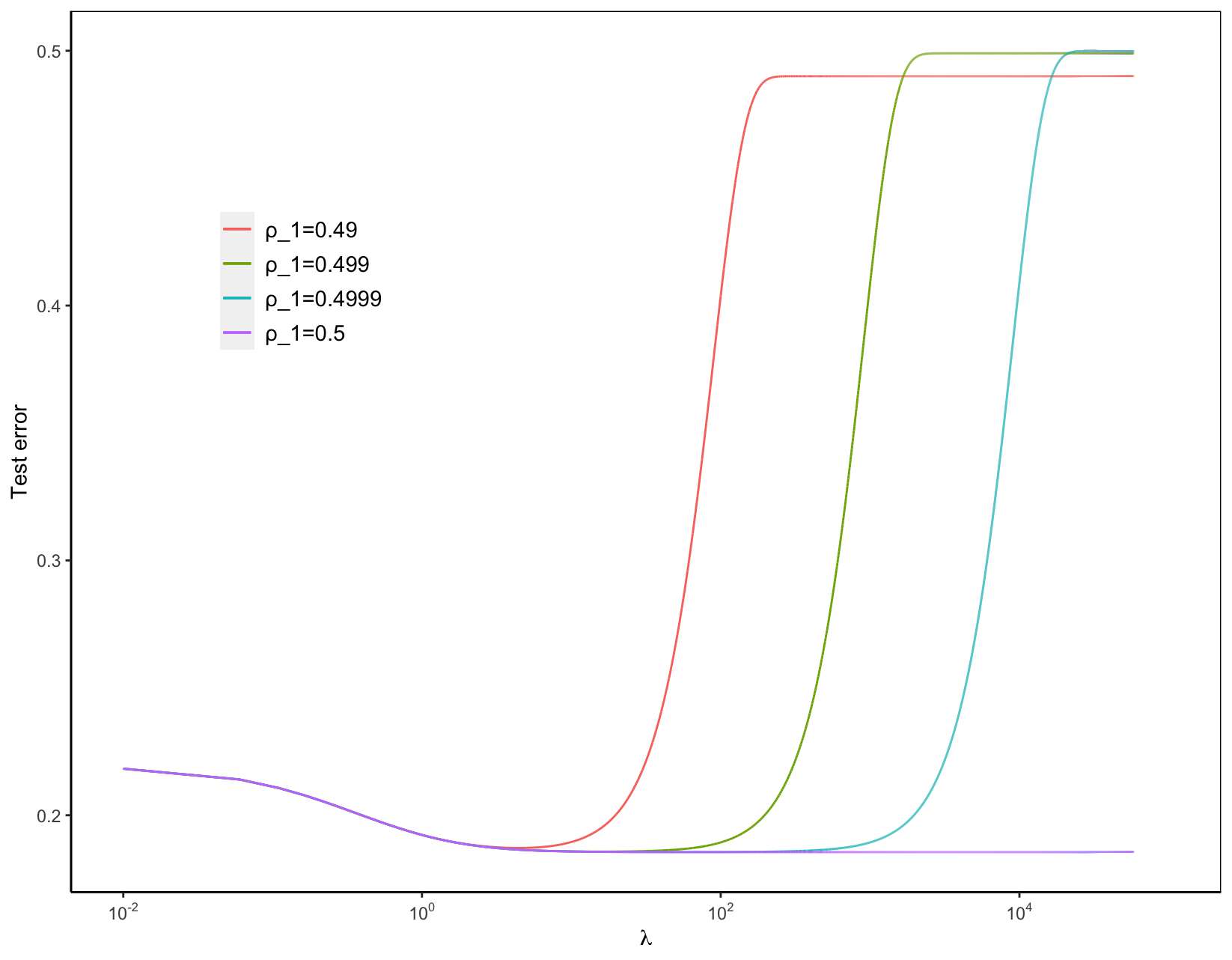}
    \end{minipage}\hfill
    \begin{minipage}{0.5\textwidth}
        \centering
        \includegraphics[width=1.\textwidth]{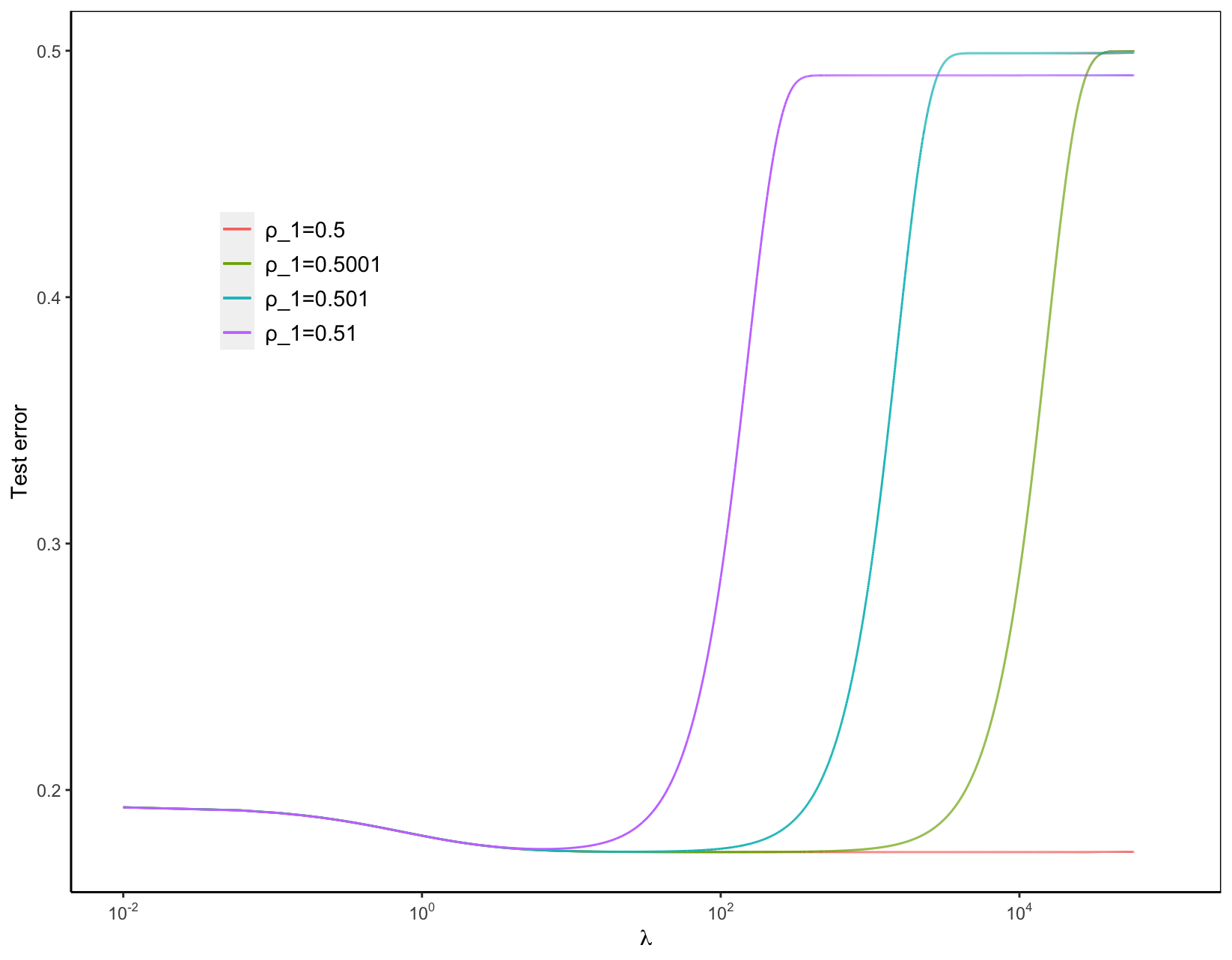}
    \end{minipage}
    \caption{Test error of model (\ref{student}) as a function of regularization. We fix the values $\alpha=4$ (left) and $\alpha=7$ (right) and consider different values of $\rho_1$ that are getting closer to 0.5.}
    \label{fig0}
\end{figure}

Here we fix $\alpha$ and view the generalization error as a function of $\lambda$. In Figure \ref{fig2} we noticed that when $\rho_1=0.5$, the generalization error is not sensitive to higher regularization. Hence we consider the cluster sizes close to $0.5$ to observe the change in the test error when regularization increases. According to Figure \ref{fig0}, for each $\rho_1$ close to $0.5$, the generalization error decreases and increases again to reach a plateau. The minimum possible test error is achieved at some finite regularization value ($\lambda^*$) and when $\lambda>\lambda^*$ test error increases again. But when $\rho_1$ is exactly 0.5, the test error keeps steady at a low level after a certain $\lambda^*$.

\section{Conclusion}\label{summ}

In this paper we work on ratio-wise double descent behavior for a two layer neural network classification model. The test error is a function of the ratio $\alpha$ between the sample size $n$ and the model dimension $d$ and we consider asymptotics of the test error when $n, d\to\infty$. We derive the exact theoretical test error of the model in Theorem \ref{The1}. Next, we perform empirical risk minimization procedure to find the unknown quantities in the test error formula. An upper bound on the local training loss is used as the candidate to observe the behavior of the asymptotics of unknown quantities. We outline these findings in  Theorem \ref{mainT}.

Utilizing the results from Theorems \ref{The1} and \ref{mainT} and using the square loss function, we plot the test error curve as a function of $\alpha$ and observe the double descent behavior when the regularization is very low. The curve's peak happens when the sample size equals the model dimension. We also notice that when the regularization increases, the peak of the curve disappears and the test error decreases monotonically. When the cluster sizes are equal, the effect of strong regularization is not significant, while with uneven cluster sizes, under strong regularization, the test error is steady at first, and then it starts decreasing as a higher value of $\alpha$ is reached. 

We confirm the existence of the double descent phenomenon in the test error for two-layer neural network model with low level of regularization. In this case, our theoretical results confirm that  when the test error is analyzed ratio-wise, the best performance of the model is achieved after the peak of the test error in the under-parameterized region. We then analyze the effect of regularization on the double descent curve. We suggest to use a suitable level of regularization in the empirical risk to have ideal test error for different ratios $\alpha$, no matter whether the cluster sizes are even or not. With this optimal test error, the double descent phenomenon disappears. Instead, the test error decreases monotonically and it is consistent with the classical idea that more data is always better. 

Here we used $l_2$ regularization to support the fact that optimal regularization can mitigate the double descent in binary classification models. In future research projects we plan to investigate the double descent phenomenon with other regularization methods like Lasso or elastic-net, which is a combination of both $l_2$ and Lasso. \\

\noindent{\bf Supplementary Material}\\
We have included two supplementary files where Supplementary material 1 contains detailed
calculations, theorems and proofs and Supplementary material 2 contains the R/RStudio codes
used to draw the curves presented in the paper.\\

\noindent{\bf Funding}\\
The research of Hailin Sang is partially supported by the Simons Foundation Grant 586789, USA.

\newpage

\section{Supplementary material 1}
\subsection{Legendre transformation}\label{leg}
Given a function $l: \mathbb{R}\to\mathbb{R}$, its \textit{Legendre transform} $\tilde l$ (the ``conjugate'' function ) is defined by
 \begin{equation}
 \label{ltilda}
\tilde l(y)=\max_{x\in\mathbb{R}}\left\{xy-l(x)\right\}.
\end{equation}

Then the Legendre transformation transforms the pair $(x, l(x))$ into a new pair $(y,\tilde l(y))$ by the definition. The domain of $\tilde l$ is the set of $y\in\mathbb{R}$ such that the supremum is finite. $y$ is known as the conjugate variable. If $l(x)$ is a convex function, the inverse transformation gives $l(x)$ back and 
\begin{equation}\label{invse}
l(x)=\max_{y\in\mathbb{R}}\left\{xy-\tilde l(y)\right\}.
\end{equation}

\noindent In our study, we replace the convex loss function $l$ by the inverse transformation of the Legendre transformation $\tilde l$ (\ref{invse}). For example, the square loss function $l(x)=\frac{1}{2}(1-x)^2$ has the conjugate given by, $\tilde l(y)=\max_{x\in\mathbb{R}}\left\{xy-\frac{1}{2}(1-x)^2\right\}=\frac{y^2}{2}+y$. This transformation allows us to convert the original minimization problem to a min-max problem.\\
\subsection{Properties of Legendre transformation}\label{legpro}
The necessary condition for the existence of $\tilde{l}(y)$ is that the derivative of the function inside the maximum in (\ref{ltilda}) with respect to $x$ is zero, i.e.,
 \begin{equation}\label{nmax}
y-l^{\prime}(x)=0 \Rightarrow l^{\prime}(x)=y.
\end{equation}
This is to be viewed as an equation of $x$ for a given $y$. 
Moreover, when the chosen $l(x)$ is a strictly convex function, the second derivative of $xy-l(x)$  with respect to $x$ is $-l^{\prime\prime}(x)$, which is negative by assumption. Therefore $l^{\prime}(x)=y$ is necessary and sufficient for the local maximum. 
It is possible that the equation (\ref{nmax}) has multiple solutions.  However, the solution is unique  if $l$ satisfies the two conditions that $l^{\prime}(x)$ is continuous and monotonically increasing and $l^{\prime}(x) \to\infty$ for $x\to\infty$ and $l^{\prime}(x)\to-\infty$ for $x\to -\infty$. Thus, under these conditions, we have an equivalent way to write $\tilde l$ via the two equations
 \begin{equation}\label{propl}
\tilde l(y)=xy-l(x) \;\;\;\; \text{ and } \;\;\;\; y=l^{\prime}(x).
\end{equation}
This can also be reduced to, $\tilde l(y)=xl^{\prime}(x)-l(x), $ provided that $y=l^{\prime}(x)$ should be solved for $x$ in terms of $y$. The differential of $\tilde l(y)=xy-l(x) $ can be written as, 
 \begin{equation*}
 \begin{split}
d \tilde l(y)&=y dx+x dy-l^{\prime}(x)dx=y dx+x dy-ydx =xdy \;\; \Rightarrow\;\; \tilde l^{\prime}(y)=x.
\end{split}
\end{equation*}

\noindent In conclusion, when the function $l$ is strictly convex and satisfies the two conditions, for $ x=\tilde l^{\prime}(y)$, we have the relationship used in (\ref{scalar change}) as $l^{\prime}(x)=y$ and $l(x)=y\tilde l^{\prime}(y)-\tilde l(y)$.
\subsection{Strategy of using Convex Gaussian Min-Max Theorem}\label{stat}
The expression in (\ref{omega v d}) is a lower bound for the auxiliary problem (\ref{ao1}). Since we are interested in high dimensional behaviors when $n, d \to \infty$, we do not need to compute these lower bounds for the auxiliary problem or local and global losses for the primary problem exactly. Instead, we use the relationships introduced in CGMT to show that, $\omega_{\lambda}^{(d)}(r,s)$ is a candidate to observe the long-run behavior of the global training loss introduced in (\ref{global training loss}). 
\begin{theorem}\label{converge}
In higher dimension when $n,d\to \infty$, the global training loss $L_{\lambda}^*$ can be approximated by the infimum of the lower bound of the local training loss $\omega_{\lambda}^{(d)}$ in auxiliary optimization problem, i.e.,
\begin{equation}\label{ucgmt}
\mathbb{P}\left( \lim_{n,d \to\infty} L^*_\lambda(r,s) = \lim_{n,d \to\infty} \omega_{\lambda}^{*}(r,s)\right)=1.
\end{equation} 

\begin{proof}
For fixed $r$ and $s$, we previously defined the local training loss $L_{\lambda}(r, s)$ in (\ref{min problem}). We set the global training loss in (\ref{global training loss}). Using AO problem, in the local training loss minimization procedure, we have found a lower bound $\omega_{\lambda}^{(d)}(r,s)$ as in (\ref{omega v d}) such that $\tilde L_{\lambda}(r,s)\geq \omega_{\lambda}^{(d)}(r,s)$. Next we define,
\begin{equation}
\label{global training loss conv}
\begin{split}
\omega_{\lambda}^*(r,s):=\inf_{s^2\leq r} \quad & \omega^{(d)}_{\lambda}(r,s).
\end{split}
\end{equation}
The first statement of CGMT resulted in the following inequality in (\ref{cgmt 3}).
 \begin{equation*}
\mathbb{P}(L_\lambda(r,s)<c) \leq 2\mathbb{P} ( \omega_{\lambda}^{(d)}(r,s)<c ).
 \end{equation*}
 By (\ref{cgmt 3}), for any $\delta>0, \mathbb{P}( L_\lambda(r,s)<\omega^{(d)}_{\lambda}(r,s)-\delta) \leq 2\mathbb{P} (\omega_{\lambda}^{(d)}(r,s)<\omega^{(d)}_{\lambda}(r,s)-\delta ).$ The right side of the inequality becomes zero since $\delta>0$ and it implies,
  \begin{align*}
 \mathbb{P}( L_\lambda(r,s)\geq \omega^{(d)}_{\lambda}(r,s)-\delta)=1.
 \end{align*}
 Then using (\ref{global training loss}) and (\ref{global training loss conv}), we can rewrite the above as,
 \begin{align*}
 \mathbb{P}( L_\lambda^*(r,s)\geq \omega_{\lambda}^*(r,s)-\delta) =1, \;\;\text{ and }\;\; \mathbb{P}( L_\lambda^*(r,s)-\omega_{\lambda}^*(r,s) \geq -\delta) =1.
  \end{align*}
 Since our interest is on the high dimensional behavior of the loss, next we consider the limits when $n, d \to \infty$. 
  \begin{align}
 \label{gside}
 \mathbb{P}\Big( \lim_{n, d\to\infty} L_\lambda^*(r,s) - \lim_{n, d\to\infty}  \omega_{\lambda}^*(r,s)\geq - \delta\Big) =1.
  \end{align} 
 
Recall the $\psi(\bm\beta,\textbf{u})$ in (\ref{si}). When $u_iy_i>0$, then $\psi(\bm{\beta},\textbf{u})$ is convex with respect to $\bm{\beta}$ due to its absolute-valued term with positive $u_iy_i$ and $\psi(\bm{\beta},\textbf{u})$ is concave with respect to $u_i$ since $-\tilde l(u_i)$ is concave by its definition (Supplementary material 1 \ref{ltilda}). Hence $\psi(\bm{\beta},\textbf{u})$ is convex-concave on $\mathbb{R}^d \times \mathbb{R}^n$ where $\bm{\beta}\in\mathbb{R}^d$ and $\textbf{u}\in\mathbb{R}^n$. Using the convex-concave property of $\psi(\bm\beta,u)$ \cite{TOH}, for any $c\in\mathbb{R}$, we have 
  \begin{equation*}
 \mathbb{P}(L_\lambda(r,s)\geq c) \leq 2\mathbb{P} ( \tilde L_{\lambda}(r,s)\geq c ).
 \end{equation*}
 Let $c=\omega_{\lambda}^{(d)}(r,s)+\delta$ for any $\delta>0$ and it yields,
   \begin{equation*}
 \mathbb{P}(L_\lambda(r,s)\geq \omega_{\lambda}^{(d)}(r,s)+\delta) \leq 2\mathbb{P} ( \tilde L_{\lambda}(r,s)\geq \omega_{\lambda}^{(d)}(r,s)+\delta ).
 \end{equation*} 
 Then using (\ref{global training loss}) and (\ref{global training loss conv}), we claim that the infimum of $L_\lambda(r,s)$ is greater than the infimum of $\omega_{\lambda}^{(d)}(r,s)+\delta$, since $L_\lambda(r,s)\geq \omega_{\lambda}^{(d)}(r,s)+\delta$. Same argument follows for the right side of the inequality above and we get, $ \mathbb{P}(L_\lambda^*(r,s)\geq \omega_{\lambda}^*(r,s)+\delta) \leq 2\mathbb{P} ( \tilde L_{\lambda}^*(r,s)\geq \omega_{\lambda}^*(r,s)+\delta )$. We rewrite this as,
    \begin{equation*}
  \mathbb{P}(L_\lambda^*(r,s)- \omega_{\lambda}^*(r,s)\geq \delta) \leq 2\mathbb{P} ( \tilde L_{\lambda}^*(r,s)- \omega_{\lambda}^*(r,s)\geq \delta ).
 \end{equation*} 
 If we consider the high dimensional behavior when $n,d \to\infty,$
    \begin{equation}
    \label{to 0}
  \mathbb{P}( \lim_{n,d \to\infty} L_\lambda^*(r,s)- \lim_{n,d \to\infty}\omega_{\lambda}^*(r,s)\geq \delta) \leq 2\mathbb{P} (\lim_{n,d \to\infty} \tilde L_{\lambda}^*(r,s)- \lim_{n,d \to\infty} \omega_{\lambda}^*(r,s)\geq \delta ).
 \end{equation} 
 By the argument shown in supplementary material Section \ref{to zero} for square loss function, we have that, 
\begin{equation*}
 \lim_{n,d\to \infty}  |\inf_{s^2\leq r} \tilde L_\lambda(r,s)- \inf_{s^2\leq r} \omega^{(d)}_{\lambda}(r,s)| \to 0.
 \end{equation*} 
Hence the right side of the (\ref{to 0}), becomes zero and, $ \mathbb{P}( \lim_{n,d \to\infty} L_\lambda^*(r,s)- \lim_{n,d \to\infty}\omega_{\lambda}^*(r,s)\geq \delta) = 0 $. This implies, $  \mathbb{P}( \lim_{n,d \to\infty} L_\lambda^*(r,s)- \lim_{n,d \to\infty}\omega_{\lambda}^*(r,s)\leq \delta) = 1.$ Combining (\ref{gside}) with the above final result, we get \ref{ucgmt}.
\end{proof}
\end{theorem}
Hence to observe the asymptotic behavior of the global training loss, we use $\omega^{(d)}_{\lambda}(r,s)$ as a candidate. Since we already have an expression for $\omega^{(d)}_{\lambda}(r,s)$ in (\ref{omega v d}), first we minimize it to find  $\omega^*_{\lambda}(r,s)$  and finally consider the high dimensional behavior by sending $n, d \to \infty$.

\subsection{ $\lim_{n,d\to \infty}  |\inf_{s^2\leq r} \tilde L_\lambda(r,s)- \inf_{s^2\leq r} \omega^{(d)}_{\lambda}(r,s)|= 0$}\label{to zero}
$\omega_{\lambda}^{(d)}(r,s)$ can be written as $ \omega_{\lambda}^{(d)}(r,s)=\frac{\lambda r}{2}+\frac{1}{d}\sum_{i=1}^{n}\frac{(1-v_i)^2}{2}$ with the square loss function $l(v_i)=\frac{(1-v_i)^2}{2}$. Then, 
\begin{equation}
\label{wsq}
\frac{\sum_{i=1}^n (v_i-1)^2}{d}=\frac{1}{d}\left(\sum_{i=1}^n v_i^2-2 \sum_{i=1}^n v_i+\sum_{i=1}^n 1\right).
\end{equation}
Note that ${\sum_{i=1}^n 1}/d=\alpha$ and $\frac{2\sum_{i=1}^n v_i}{d}=\frac{2\sum_{i=1}^n y_i\sigma(\frac{\textbf{x}_i^T\bm{\beta}}{\sqrt{d}}+b)}{d}$, where
\begin{equation*}
\begin{split}
\frac{\sum_{i=1}^n \sigma(\frac{\textbf{x}_i^T\bm{\beta}}{\sqrt{d}}+b)}{d}\leq   \frac{\sum_{i=1}^n |\frac{\textbf{x}_i^T\bm{\beta}}{\sqrt{d}}+b|}{d}=\frac{\sum_{i=1}^n |\textbf{x}_i^T\bm{\beta}+b\sqrt{d}|}{d^{1.5}}\leq \frac{\sum_{i=1}^n |\textbf{x}_i^T\bm{\beta}|}{d^{1.5}}+\alpha|b|
\end{split}
\end{equation*}
For the first term we substitute the teacher model in (\ref{teacher}), 
\begin{equation*}
\begin{split}
 \frac{\sum_{i=1}^n |\textbf{x}_i^T\bm{\beta}|}{d^{1.5}}&= \frac{1}{d^{1.5}}\sum_{i=1}^n \left | \left(\frac{\bm\eta y_i}{\sqrt{d}}+\epsilon_i\right)^T\bm\beta \right |=\frac{1}{d^{1.5}} \sum_{i=1}^n \left | \frac{y_i\bm\eta^T\bm\beta}{\sqrt{d}}+\epsilon_i^T\bm\beta \right |\\
& \leq \frac{s}{d} \sum_{i=1}^n | y_i | +  \frac{1}{d^{1.5}} \sum_{i=1}^n | \epsilon_i^T\bm\beta |=s\alpha+\frac{1}{d^{1.5}} \sum_{i=1}^n | \epsilon_i^T\bm\beta | \to s\alpha+\alpha\sqrt{\frac{2r}{\pi}}
\end{split}
\end{equation*}
by the law of large numbers. Next we work on the third term in (\ref{wsq}).
\begin{equation*}
\begin{split}
 \frac{\sum_{i=1}^n v_i^2}{d}&=\frac{\sum_{i=1}^n (y_i\sigma(\frac{\textbf{x}_i^T\bm{\beta}}{\sqrt{d}}+b))^2}{d}\leq \frac{\sum_{i=1}^n(\textbf{x}_i^T\bm\beta+b\sqrt{d})^2}{d^2} \leq =2\frac{\sum_{i=1}^n(\textbf{x}_i^T\bm\beta)^2}{d^2}+2\alpha b^2.
 \end{split}
\end{equation*}
The first term can be simplified using the teacher model as,
\begin{equation*}
\begin{split}
\frac{\sum_{i=1}^n(\textbf{x}_i^T\bm\beta)^2}{d^2}&= \frac{1}{d^2}\sum_{i=1}^n \left(\left(\frac{\bm\eta y_i}{\sqrt{d}}+\epsilon_i\right)^T\bm\beta \right)^2 \leq \frac{2}{d^2}\sum_{i=1}^n \left(\frac{y_i\bm\eta^T\bm\beta}{\sqrt{d}}\right) ^2 +\frac{2}{d^2}\sum_{i=1}^n \left(\epsilon_i^T\bm\beta  \right)^2\\
&=\frac{2s^2\alpha}{d}+\frac{2}{d^2}\sum_{i=1}^n \left(\epsilon_i^T\bm\beta  \right)^2 \to 2r\alpha
\end{split}
\end{equation*}

Thus the third term in (\ref{wsq}) goes to $2r\alpha$ when $d\to\infty$ for a fixed value of $\alpha$. Hence $\omega^{(d)}_{\lambda}(r,s)$ is bounded. This, in particular, implies that 
\begin{equation}\label{eqnum}
| \inf_{s^2\leq r} \tilde L_\lambda(r,s)- \inf_{s^2\leq r} \omega^{(d)}_{\lambda}(r,s)|  \leq \sup_{s^2\leq r}|\tilde L_{\lambda}(r,s)-\omega^{(d)}_{\lambda}(r,s)|.
\end{equation}
Indeed, if both $\tilde L_\lambda(r,s) $ and $\omega^{(d)}_{\lambda}(r,s)$ are bounded functions, then the above inequality follows, and in case only $\omega^{(d)}_{\lambda}(r,s)$ is bounded, then the right side of \eqref{eqnum} is infinite. Taking the limits $n,d\to \infty$ in \eqref{eqnum}  gives
\begin{equation*}
\begin{split}
&\lim_{n,d\to \infty} | \inf_{s^2\leq r} \tilde L_\lambda(r,s)- \inf_{s^2\leq r} \omega^{(d)}_{\lambda}(r,s)|  \leq \lim_{n,d\to\infty} \sup_{s^2\leq r}|\tilde L_{\lambda}(r,s)-\omega^{(d)}_{\lambda}(r,s)|\\
&=\lim_{n,d\to\infty} \sup_{s^2\leq r} \Big |\max_{\substack{\textbf{u},  \\ u_iy_i>0}}\;\Big\{\frac{1}{d}\sum_{i=1}^{n}\Big(\frac{u_{i}(s+y_ib+\sqrt{r}g_i)}{2}-\tilde l(u_i)\Big)\\
 & +\min_{\bm{\beta}} \; \Big\{\frac{||\textbf{u}||_2\textbf{h}^T\bm{\beta}}{2d\sqrt{d}}+\frac{1}{d}\sum_{i=1}^{n}\frac{u_iy_i}{2\sqrt{d}}|\textbf{x}_i^T\bm{\beta}+b\sqrt{d}|\Big\}\Big\}-\max_{\substack{\textbf{u},  \\ u_iy_i>0}}\Big\{\frac{1}{d}\sum_{i=1}^{n}\Big(\frac{u_{i}(s+y_ib+\sqrt{r}g_i)}{2}-\tilde l(u_i)\Big)\\
 & \quad +\Big(\frac{s\bm{\eta}^T\textbf{h}}{d\sqrt{d}}-\sqrt{\frac{r-s^2}{d}}\Big)\frac{||\textbf{u}||_2}{2} +\min_{\bm{\beta}} \; \frac{1}{d}\sum_{i=1}^{n}\frac{u_iy_i}{2\sqrt{d}}|\textbf{x}_i^T\bm{\beta}+b\sqrt{d}|\Big\}\Big |
\end{split}
\end{equation*}
Simplifying further,
\begin{equation*}
\begin{split}
&\lim_{n,d\to \infty} | \inf_{s^2\leq r} \tilde L_\lambda(r,s)- \inf_{s^2\leq r} \omega^{(d)}_{\lambda}(r,s)| \\
& \quad  \leq \lim_{n,d\to\infty} \sup_{s^2\leq r} \;\max_{\substack{\textbf{u},  \\ u_iy_i>0}}\;\Big |\min_{\bm{\beta}} \; \Big\{\frac{||\textbf{u}||_2\textbf{h}^T\bm{\beta}}{2d\sqrt{d}}+\frac{1}{d}\sum_{i=1}^{n}\frac{u_iy_i}{2\sqrt{d}}|\textbf{x}_i^T\bm{\beta}+b\sqrt{d}|\Big\}\\
& \quad - \Big(\frac{s\bm{\eta}^T\textbf{h}}{d\sqrt{d}}-\sqrt{\frac{r-s^2}{d}}\Big)\frac{||\textbf{u}||_2}{2} -\min_{\bm{\beta}} \; \frac{1}{d}\sum_{i=1}^{n}\frac{u_iy_i}{2\sqrt{d}}|\textbf{x}_i^T\bm{\beta}+b\sqrt{d}| \Big|\\
&= \lim_{n,d\to\infty} \sup_{s^2\leq r} \;\max_{\substack{\textbf{u},  \\ u_iy_i>0}}\;\Big |\min_{\bm{\beta}} \; \Big\{\frac{||\textbf{u}||_2\textbf{h}^T\bm{\beta}}{2d\sqrt{d}}+\frac{1}{d}\sum_{i=1}^{n}\frac{u_iy_i}{2\sqrt{d}}|\textbf{x}_i^T\bm{\beta}+b\sqrt{d}|\Big\}-\Big(\frac{s\bm{\eta}^T\textbf{h}}{d\sqrt{d}}-\sqrt{\frac{r-s^2}{d}}\Big)\frac{||\textbf{u}||_2}{2} \Big|.
\end{split}
\end{equation*}

We see that when $n, d \to \infty$, the last expression reaches 0 given that $||\textbf{u}||_2/d$ is bounded. The boundedness of $||\textbf{u}||_2/d$ follows from boundedness of $\omega_{\lambda}^{(d)}(r,s)$ since
\begin{equation*}
\label{unorm}
\frac{||\textbf{u}||_2^2}{d^{2}}= \frac{\sum_{i=1}^n (v_i-1)^2}{d^{2}}\leq\omega_{\lambda}^{(d)}(r,s).
\end{equation*}

Thus, $\lim_{n,d\to \infty}  |\inf_{s^2\leq r} \tilde L_\lambda(r,s)- \inf_{s^2\leq r} \omega^{(d)}_{\lambda}(r,s)|\to 0$ under the square loss. 
\subsection{Finding derivatives of (\ref{omega v d}) with respect to $r,s$, and $b$ }\label{derivatives}
\ \\
\emph{\textbf{Derivative with respect to $r$}}\\

Differentiating (\ref{omega v d}) with respect to $r$, and setting it to be zero, we have
\begin{equation}
\label{33r}
\frac{d \omega^{(d)}_{\lambda}(r,s)}{d r}=\frac{\lambda}{2}+\frac{\alpha}{n}\sum_{i=1}^{n}l^{\prime}(v_{i})\frac{d v_i}{d r} =0 \Longrightarrow  \sum_{i=1}^{n}l^{\prime}(v_{i})\frac{d v_i}{d r} = -\frac{\lambda n}{2\alpha}.
\end{equation}
Differentiating (\ref{rel 1}) with respect to $r$ yields, $\gamma l^{\prime\prime}(v_i)\frac{dv_i}{dr}+l^{\prime}(v_i)\frac{d\gamma}{dr}+\frac{dv_i}{dr}=\frac{g_i}{4\sqrt{r}}. $
Rearranging the terms will yield,
\begin{equation}
\label{31r}
l^{\prime\prime}(v_i)\frac{dv_i}{d r}= \frac{1}{\gamma} \left(\frac{g_i}{4\sqrt{r}}-\frac{dv_i}{dr}-l^{\prime}(v_i)\frac{d\gamma}{dr} \right).
\end{equation}
Differentiating (\ref{gamma^2}) with respect to $r$, 
\begin{equation*}
8\alpha \gamma ||l^{\prime}(v)||_2^2\frac{d\gamma}{d r}+8\alpha\gamma^2 \sum_{i=1}^n l^{\prime}(v_i)l^{\prime\prime}(v_i)\frac{dv_i}{dr}=\left (\sqrt{r-s^2}-\frac{s\bm\eta^T\textbf{h}}{d}\right)\frac{n}{\sqrt{r-s^2}}.
\end{equation*}
Next we substitute (\ref{33r}) and (\ref{31r}) in the above expression. After some algebra we have  
\begin{equation}
\label{fixed r}
\frac{\alpha}{\sqrt{r}}\frac{1}{n}\sum_{i=1}^n g_il^{\prime}(v_i)=-2\lambda+\frac{1}{2\gamma}-\frac{s\bm\eta^T\textbf{h}}{2\gamma d \sqrt{r-s^2}}.
\end{equation}
For easy computation, define $w_i=(s+y_ib+\sqrt{r}g_i)$ and $w_i$ follows a normal distribution with mean $s+y_ib$ and standard deviation $\sqrt{r}$ conditioned on $y_i$ for each $1\leq i\leq n$. Now we rewrite (\ref{rel 1}) as $2v_{i} +2\gamma l^{\prime}(v_{i}) =w_i$ and obtain $l^{\prime}(v_i)$ as below for all $v_i\in \mathbb{R}$,
\begin{equation}
\label{l prime}
l^{\prime}(v_i)=\frac{w_i-2v_i}{2\gamma}.
\end{equation}
Then we update the relationship in (\ref{fixed r}) using the above substitution to get (\ref{r final}).

\ \\
\emph{\textbf{Derivative with respect to $s$}}\\

\noindent First we differentiate (\ref{omega v d}) with respect to $s$ and make it equal to 0 to have
\begin{flalign}
\label{omega s}
 \frac{d \omega^{(d)}_{\lambda}(r,s)}{d s} =\frac{\alpha}{n}\sum_{i=1}^n l^{\prime}(v_i)\frac{dv_i}{ds}=0 \Longrightarrow \sum_{i=1}^n l^{\prime}(v_i) \frac{dv_i}{ds}=0.&
\end{flalign}
 Differentiating (\ref{rel 1}) with respect to $s$ gives us, $ l^{\prime}(v_i)\frac{d \gamma}{d s}+l^{\prime\prime}(v_i)\gamma\frac{dv_i}{ds}+\frac{dv_i}{ds}=\frac{1}{2}$.  Rearranging the terms will yield,
\begin{equation}
\label{l 2 prime s}
l^{\prime\prime}(v_i)\frac{d v_i}{d s}= \frac{1}{\gamma}\left(\frac{1}{2}-\frac{dv_i}{ds}-l^{\prime}(v_i)\frac{d \gamma}{d s}\right).
\end{equation}
Differentiating (\ref{gamma^2}) with respect to $s$ gives the following expression.
\begin{flalign*}
 8\alpha \gamma ||l^{\prime}(v)||_2^2\frac{d \gamma}{d s}+8\alpha\gamma^2\sum_{i=1}^nl^{\prime}(v_i)l^{\prime\prime}(v_i)\frac{dv_i}{ds}=2n\left (\sqrt{r-s^2}-\frac{s\bm\eta^T\textbf{h}}{d}\right)\left(-\frac{s}{\sqrt{r-s^2}}-\frac{\bm\eta^T\textbf{h}}{d}\right).&
\end{flalign*}
We let $G=\sqrt{r-s^2}\frac{\bm\eta^T\textbf{h}}{d}-\frac{s^2\bm\eta^T\textbf{h}}{d\sqrt{r-s^2}}-s\left(\frac{\bm\eta^T\textbf{h}}{d}\right)^2 $ and write the above expression as,
\begin{equation*}
8\alpha\gamma||l^{\prime}(v)||_2^2\frac{d\gamma}{ds}+8\alpha\gamma^2\sum_{i=1}^nl^{\prime}(v_i)l^{\prime\prime}(v_i)\frac{dv_i}{ds}=-2n\left(s+G\right).
\end{equation*}
Simplifying the above expression using (\ref{l 2 prime s}) and (\ref{omega s}) yields, $-2\alpha\gamma \frac{1}{n}\sum_{i=1}^nl^{\prime}(v_i)=s+G$. Replacing $l^{\prime}(v_i)$ by (\ref{l prime}) results in the expression in (\ref{s final}).

\newpage


\emph{\textbf{Derivative with respect to $b$}}\\

\noindent Finally, we follow the same procedure for the bias term $b$ by differentiating (\ref{omega v d}) with respect to $b$ and make it equal to 0 to get the following relationship.
 \begin{flalign}
 \label{omega b}
 \frac{d \omega^{(d)}_{\lambda}(r,s)}{d b} =\frac{\alpha}{n}\sum_{i=1}^nl^{\prime}(v_i)\frac{dv_i}{db}=0 \Longrightarrow \sum_{i=1}^nl^{\prime}(v_i)\frac{dv_i}{db}=0. &
\end{flalign}
  Differentiating (\ref{rel 1}) with respect to $b$ gives, $ \frac{d \gamma}{db} l^{\prime}(v_i)+\gamma l^{\prime\prime}(v_i)\frac{dv_i}{db}+\frac{dv_i}{db}=\frac{y_i}{2}$. Rearranging the terms will yield,
\begin{equation}
\label{l 2 prime b}
l^{\prime\prime}(v_{i})\frac{d v_i}{d b}= \frac{1}{\gamma}\left(\frac{y_i}{2}-\frac{dv_i}{db}-l^{\prime}(v_i)\frac{d \gamma}{d b}\right).
\end{equation}
Differentiating (\ref{gamma^2}) with respect to $b$ gives the following expression.
\begin{flalign*}
 8\alpha \gamma \frac{d \gamma}{d b}||l^{\prime}(v)||_2^2+8\alpha\gamma^2\sum_{i=1}^nl^{\prime}(vi)l^{\prime\prime}(v_i)\frac{d v_i}{d b}=0.&
\end{flalign*}
Simplifying the above expression using (\ref{l 2 prime b}) and (\ref{omega b}) yields, $\frac{1}{2}\sum_{i=1}^ny_il^{\prime}(v_i)=0$. We end up with the relationship as shown in (\ref{fixed b}) after replacing $l^{\prime}(v_i)$ by (\ref{l prime}). 

\subsection{Application of Theorem \ref{mainT} for Square Loss in empirical risk minimization procedure}\label{curve draw calc}




In this section we workout the fixed point equations in Theorem \ref{mainT} for square loss $l(v_i)=\frac{1}{2}(1-v_i)^2$ and plot the curves for generalization error given in Theorem \ref{The1}. \\

For all $v_i\in \mathbb{R}$ we have $l^{\prime}(v_i)=v_i-1$, and by (\ref{t1 v}) we get
\begin{equation}
\gamma(v_i-1)+v_i=\frac{1}{2}w_i \Longrightarrow v_i=\frac{w_i+2\gamma}{2(\gamma+1)}.
\end{equation}

From this we compute 
\begin{equation}
w_i-2v_i=\frac{\gamma(w_i-2)}{\gamma+1},
\end{equation}
which we use later to simplify the equations introduced in Theorem \ref{mainT}.

Recall that $w_i=s+y_ib+\sqrt{r}g_i\sim\mathcal{N}(s+y_ib,r)$ and $y_i=\pm1$ with probabilities $\rho_1$ and $\rho_{-1}$, respectively, and $g_i\sim\mathcal{N}(0,1)$. Also, applying the law of large numbers, we simplify (\ref{t1 r}) as follows:

\begin{equation*}
\begin{split}
\lim_{n,d\to\infty} \left(\frac{1}{n}\sum_{i=1}^n g_i(w_i-2v_i)\right)&=\lim_{n,d\to\infty} \left(\frac{1}{n}\sum_{i=1}^n g_i \frac{\gamma(w_i-2)}{\gamma+1} \right)\\
&=\frac{\gamma^*}{1+\gamma^*} \lim_{n,d\to\infty} \left(\frac{1}{n}\sum_{i=1}^n g_i (w_i-2) \right)\\
&=\frac{\gamma^*}{1+\gamma^*} \lim_{n,d\to\infty} \left(\frac{1}{n}\sum_{i=1}^n g_i (s+y_ib+\sqrt{r}g_i-2) \right)\\
&=\frac{\gamma^*}{1+\gamma^*} \lim_{n,d\to\infty} \left(\frac{1}{n}\sum_{i=1}^n g_is+g_iy_ib+\sqrt{r}g_i^2-2g_i) \right)\\
&=\frac{\gamma^*}{1+\gamma^*} (0+0+\sqrt{r^*}-0)=\frac{\gamma^*\sqrt{r^*}}{1+\gamma^*}.
\end{split}
\end{equation*}

Then from (\ref{t1 r}) we get
\begin{equation*}
\frac{\alpha}{\sqrt{r^*}}\;\frac{\gamma^*\sqrt{r^*}}{1+\gamma^*}=-4\lambda\gamma^*+1.
\end{equation*}
Rearranging the terms we derive the formula for $\gamma^*$:
\begin{equation}
\label{g star}
\gamma^*=\frac{-(\alpha+4\lambda-1)\pm\sqrt{(\alpha+4\lambda-1)^2+16\lambda}}{8\lambda}.
\end{equation}

Next we compute the limits in (\ref{t1 b}):
\begin{equation*}
\begin{split}
\lim_{n,d\to\infty} \left(\frac{1}{n}\sum_{i=1}^ny_i(w_i-2v_i)\right)&=
\lim_{n,d\to\infty} \left(\frac{1}{n}\sum_{i=1}^ny_i \frac{\gamma(w_i-2)}{\gamma+1}\right)\\
&=\frac{\gamma^*}{\gamma^*+1}\lim_{n,d\to\infty} \left(\frac{1}{n}\sum_{i=1}^ny_i(w_i-2)\right)\\
&=\frac{\gamma^*}{\gamma^*+1}(s^*(2\rho_1-1)+b-2(2\rho_1-1))=0,\\
\end{split}
\end{equation*}
yielding 
\begin{equation}
\label{b1}
b=(2-s^*)(2\rho_1-1).
\end{equation}\\

Similarly, (\ref{t1 s}) can be rewritten as
\begin{equation*}
\begin{split}
\lim_{n,d\to\infty} \left(\frac{1}{n}\sum_{i=1}^n(w_i-2v_i)\right)&=\lim_{n,d\to\infty} \left(\frac{1}{n}\sum_{i=1}^n \frac{\gamma(w_i-2)}{\gamma+1} \right)\\
&=\frac{\gamma^*}{\gamma^*+1}\lim_{n,d\to\infty} \left(\frac{1}{n}\sum_{i=1}^n (w_i-2) \right)\\
&=\frac{\gamma^*}{\gamma^*+1}(s^*+b(1\cdot \rho_1+(-1)\cdot\rho_{-1})-2)\\
&=\frac{\gamma^*}{\gamma^*+1}(s^*+b(2\rho_1-1)-2).
\end{split}
\end{equation*}

Combining this with (\ref{b1}) we get
\begin{equation*}
\begin{split}
\frac{-\alpha\gamma^*}{1+\gamma^*}(s^*+b(2\rho_1-1)-2)&=s^*,\\
\frac{-\alpha\gamma^*}{1+\gamma^*}(s^*+(2-s^*)(2\rho_1-1)(2\rho_1-1)-2)&=s^*,\\
\frac{-\alpha\gamma^*}{1+\gamma^*}(s^*+(2-s^*)(2\rho_1-1)^2-2)&=s^*,\\
\frac{-\alpha\gamma^*}{1+\gamma^*}(2-s^*)((2\rho_1-1)^2-1)&=s^*,\\
\frac{4\alpha\gamma^*\rho_1\rho_{-1}}{1+\gamma^*}(2-s^*)&=s^*,
\end{split}
\end{equation*}\\
which gives
\begin{equation}
\label{s final s}
\begin{split}
s^*&=\frac{8\alpha\gamma^*\rho_1\rho_{-1}}{1+\gamma^*+4\alpha\gamma^*\rho_1\rho_{-1}}.
\end{split}
\end{equation}\\

Hence $b$ in (\ref{b1}) simplifies to
\begin{equation}
\label{b final}
 b=(2-s^*)(2\rho_1-1)= \frac{2(1+\gamma^*)(2\rho_1-1)}{1+\gamma^*+4\alpha\gamma^*\rho_1\rho_{-1}}.
\end{equation}

So far we have found $\gamma^*, s^*$ and $b$ in terms of the known quantities. Finally we do the limit computation in (\ref{t1 gamma}),
\begin{equation*}
\begin{split}
&\lim_{n,d\to\infty} \left( \frac{1}{n}\sum_{i=1}^n(w_i-2v_i)^2\right)=\lim_{n,d\to\infty} \left( \frac{1}{n}\sum_{i=1}^n \left(\frac{\gamma(w_i-2)}{\gamma+1}\right)^2\right)\\
&=\left(\frac{\gamma^*}{\gamma^*+1}\right)^2 \lim_{n,d\to\infty} \left( \frac{1}{n}\sum_{i=1}^n (w_i-2)^2\right)\\
&=\left(\frac{\gamma^*}{\gamma^*+1}\right)^2 \lim_{n,d\to\infty} \left( \frac{1}{n}\sum_{i=1}^n w_i^2-4w_i+4\right)\\
&=\left(\frac{\gamma^*}{\gamma^*+1}\right)^2 \lim_{n,d\to\infty} \left( \frac{1}{n}\sum_{i=1}^n (s+y_ib+\sqrt{r}g_i)^2-4w_i+4\right)\\
&=\left(\frac{\gamma^*}{\gamma^*+1}\right)^2 ((s^*)^2+b^2+r^*+2s^*b(2\rho_1-1)-4s^*-4b(2\rho_1-1)+4).
\end{split}
\end{equation*}

Hence (\ref{t1 gamma}) simplifies to,
\begin{equation}
\alpha \left(\frac{\gamma^*}{\gamma^*+1}\right)^2 ((s^*)^2-b^2+r^*-4(s^*-1))=r^*-(s^*)^2.
\end{equation}
Solving this for $r^*$ yields
\begin{equation}
\label{r final s}
r^*=\frac{\alpha(\gamma^*)^2((s^*-2)^2-b^2)+(\gamma^*+1)^2(s^*)^2}{(1+\gamma^*)^2-\alpha(\gamma^*)^2}.
\end{equation}

The obtained asymptotic values of $\gamma, r, s$ together with the asympotic value of the bias term $b$ are presented below (note that we always pick the positive value of $\gamma$):
\begin{equation}
\label{sqe}
\begin{split}
\gamma^*&=\frac{-(\alpha+4\lambda-1)\pm\sqrt{(\alpha+4\lambda-1)^2+16\lambda}}{8\lambda},\\
s^*&=\frac{8\alpha\gamma^*\rho_1\rho_{-1}}{1+\gamma^*+4\alpha\gamma^*\rho_1\rho_{-1}},\\
 b^*&=(2-s^*)(2\rho_1-1),\\
r^*&=\frac{\alpha(\gamma^*)^2((s^*-2)^2-b^2)+(\gamma^*+1)^2(s^*)^2}{(1+\gamma^*)^2-\alpha(\gamma^*)^2}.
\end{split}
\end{equation}
Recall that from Theorem \ref{The1}
 \begin{equation}
\label{test error*}
R^*(\bm{\hat\beta})=1-\rho_{1}\Phi\left(\frac{s^*+b^*}{\sqrt{r^*}}\right)-\rho_{-1}\Phi\left(\frac{s^*-b^*}{\sqrt{r^*}}\right),
\end{equation}
and the values of $r^*, s^*$ and $ b^*$ are given in (\ref{sqe}). \\

\newpage
\section{Supplementary material 2}
\subsection{Generating the plots in Section \ref{viz} and Section \ref{curve sq}  }\label{curve draw}

Here we present the R/RStudio code used to draw the figures in Section \ref{viz} and Section \ref{curve sq}.

\begin{lstlisting}[language=R]
#---------------------------------------------------------------
# Wisconsin Breast Cancer Dataset Analysis in Section 2
#---------------------------------------------------------------
library(tidyverse)
library(keras)
library(caret)
library(fastDummies)
library(readxl)
library(dplyr)
library(tensorflow)  

set.seed(42)
tensorflow::set_random_seed(42)  # Ensures reproducibility in Keras

# Load data
df <- read_excel("wisconsin_data.xlsx")
df$Y <- ifelse(df$Y == "B", -1, 1)  # Convert 'B' to -1 and 'M' to 1
final <- df

# Normalize the features
X <- final %>% 
  select(-Y) %>% 
  scale()

y <- to_categorical(final$Y)

# Function to train the model and return test error
train_and_evaluate <- function(X, y, sample_size) {
  # Ensure the sample size doesn't exceed the dataset size
  sample_size <- min(sample_size, nrow(X))
  
  # Subset data to the current sample size
  set.seed(42)  # Ensure reproducibility for random sampling
  sample_indices <- sample(1:nrow(X), size = sample_size, replace = FALSE)
  
  X_train_subset <- X[sample_indices, , drop = FALSE]
  y_train_subset <- y[sample_indices, ]
  
  # Define the test set as the rest of the data
  X_test <- X[-sample_indices, , drop = FALSE]
  y_test <- y[-sample_indices, ]
  
  # Build the model
  model <- keras_model_sequential() %>% 
    layer_dense(units = 30, activation = 'relu', input_shape = ncol(X), 
    kernel_regularizer = regularizer_l2(0.000001)) %>% 
    layer_dropout(rate = 0.3) %>% 
    layer_dense(units = 2, activation = 'sigmoid')
  
  model %>% compile(
    loss = 'binary_crossentropy',
    optimizer = 'adam',
    metrics = c('accuracy')
  )
  
  # Train the model
  model %>% fit(
    X_train_subset, y_train_subset, 
    epochs = 200, 
    batch_size = 5,
    validation_data = list(X_test, y_test),
    verbose = 0
  )
  
  # Evaluate the model on the test data
  test_results <- model %>% evaluate(X_test, y_test, verbose = 0)
  
  # Return the loss (test error)
  return(test_results[1])  # Access loss directly
}

# Sample sizes to consider (increasing from 5 to the size of the full dataset)
sample_sizes <- c(seq(10, 40, 5), seq(50, 550, by = 10))

# Record test errors for each sample size
test_errors <- sapply(sample_sizes, function(n) {
  train_and_evaluate(X, y, n)
})

# Create a data frame for plotting
results <- data.frame(
  Alpha = sample_sizes / 30,
  TestError = test_errors
)

require(mass)
library(scales)

# Plot the test error curve
ggplot(results, aes(x = Alpha, y = TestError, color = "lambda = 0.000001")) +  
  geom_line() +
  geom_point() +
  labs(
    title = "Test Error vs. Sample Size/Dimension",
    x = expression(paste(alpha)),
    y = "Test Error"
  ) +
  scale_x_continuous(n.breaks = 8, limits = c(0, 8)) +  
  geom_vline(xintercept = 1, linetype = "dashed", color = "black", size = 0.5) +  # Add vertical line at x = 1
  scale_color_manual(values = c("lambda = 0.000001" = "blue")) +  
  theme(
    panel.grid.major = element_blank(),
    panel.grid.minor = element_blank(),
    panel.background = element_blank(),
    axis.line = element_line(colour = "black"),
    panel.border = element_rect(color = "black", fill = NA, size = 0.5),
    legend.text = element_text(size = rel(1)), 
    legend.title = element_blank()
  )
\end{lstlisting} \ \\

\begin{lstlisting}[language=R]
#---------------------------------------------------------------
# Generating plots in Section 6
#---------------------------------------------------------------
library(tidyverse)
library(matlib)
library(MASS)
library(quadprog)
library("ggplot2")
set.seed(100)

quad <- function(a, b, c)
{
  a <- as.complex(a)
  answer <- c((-b + sqrt(b^2 - 4 * a * c)) / (2 * a),
              (-b - sqrt(b^2 - 4 * a * c)) / (2 * a))
  if(all(Im(answer) == 0)) answer <- Re(answer)
  if(answer[1] == answer[2]) return(answer[1])
  if(answer[1]<=0)return(answer[2])
  if(answer[2]<=0)return(answer[1])
  answer
}

# Parameters
features <- 60 # Dimension/ number of features
N<-300 # maximum sample size
eta <- rnorm(features)  # Gaussian vector eta
l2_lambda <- 1e-5 # L2 regularization 
alpha<-N/features # Ratio of sample size/features, increase alpha from 1 to 10

y <- sample(c(1, -1), size = N, replace = TRUE) # Random binary vector of length N
probabilities <- table(y) / sum(table(y))
rho1<-probabilities[[2]] # probability of getting y=-1
rho2<-probabilities[[1]] # probability of getting y=1
x <- matrix(0, nrow = N, ncol = features) # Initialize matrix for x_i
for (i in 1:N) {
  epsilon <- rnorm(features)   # Generate epsilon 
  x <- (eta * y)/sqrt(features) + epsilon # Calculate x_i
}

steps<-0.05 # Steps to plot the curve
TestError<-matrix(data=NA, ncol=1, nrow=(alpha/steps))
begin<-0.05
i=0
j=0
for(l in c(l2_lambda)){
  j<-j+1
  for(a in seq(begin,alpha,steps)){
    i<-i+1
    g<-quad(4*l,a+4*l-1,-1)
    g1<-sqrt((a+4*l-1)^2+16*l)
    s<-(8*a*g*rho1*rho2)/(1+g+4*a*rho1*rho2*g)
    b<-(2*rho1-1)*(2-s)
    r<-((a*g^2*(-b^2+(s-2)^2)+(1+g)^2*s^2)/((1+g)^2-a*g^2))
    TestError[i,j]<-1-rho1*pnorm((s+b)/sqrt(r),0,1)-rho2*pnorm((s-b)/sqrt(r),0,1)
    print(r)
  }

  i=0
  
}

# Reshape data frame
a = seq(begin,alpha,steps)
l1=TestError[,1]

k<-which.max(TestError[,1])
ggplot() + geom_line(aes(x=a,y=l1,group=1, color='\u03BB=0.00001')) +
  labs(x=expression(paste(alpha)),y="Test error")+
  scale_x_continuous(breaks = c(seq(0, 10, by = 1)))+
  theme(legend.position = c(0.8, 0.6),legend.title = element_blank())+
  geom_vline(xintercept = k*steps, linetype="dotted")+
  theme(panel.grid.major = element_blank(), panel.grid.minor = element_blank(),
        panel.background = element_blank(), 
        axis.line = element_line(colour = "black"),
        panel.border = element_rect(color = "black", fill = NA,
                                    size = 0.5))+
  theme(legend.text=element_text(size=rel(1)))



\end{lstlisting}



\begin{thebibliography}{99} 

\bibitem{AAH} Advani, M. S., Andrew, M. S., and Haim, S. (2020). High-dimensional dynamics of generalization error in neural networks. {\it Neural Networks, 132}, 428-446.


\bibitem{AKL} Amir, I., Koren, T., and Livni, R. (2021). SGD generalizes better than GD (and regularization doesn’t help). {\it Conference on Learning Theory}, 63-92.


\bibitem{BHMM} Belkin, M., Hsu, D., Ma, S. and Mandal, S. (2019). Reconciling modern machine-learning practice and the classical bias–variance trade-off. {\it Proceedings of the National Academy of Sciences, 116}({\bf32}), 15849-15854.

\bibitem{BG} Bhavsar, H., and Ganatra, A. (2012). A comparative study of training algorithms for supervised machine learning. {\it International Journal of Soft Computing and Engineering (IJSCE), 2}({\bf04}), 2231-2307.

\bibitem{BG2} Bonaccorso, G. (2017). {\it Machine Learning Algorithms}. Packt Publishing Ltd. Birmingham, UK.




\bibitem{DRBK} D’Ascoli, S., Refinetti, M., Biroli, G., and Krzakala, F. (2020). Double trouble in double descent: Bias and variance (s) in the lazy regime. {\it International Conference on Machine Learning}, 2280-2290.


\bibitem{DKT} Deng, Z., Kammoun, A., and Thrampoulidis, C. (2022). A model of double descent for high-dimensional binary linear classification. {\it Information and Inference: A Journal of the IMA, 11}({\bf02}), 435-495.


\bibitem{GJSGS+} Geiger, M., Jacot, A., Spigler, S., Gabriel, F., Sagun, L., d’Ascoli, S., Biroli, G., Hongler, C., and  Wyart, M. (2020). Scaling description of generalization with number of parameters in deep learning. {\it Journal of Statistical Mechanics: Theory and Experiment}, ({\bf02}), 023401.


\bibitem{HKL} Hutter, F., Kotthoff, L., and Vanschoren, J. (2019). Automated machine learning: methods, systems, challenges. {\it Springer Nature}, 219.




\bibitem{KT} Kini, G. R. and Thrampoulidis, C. (2020). Analytic study of double descent in binary classification: The impact of loss. {\it 2020 IEEE International Symposium on Information Theory (ISIT)}, 2527-2532.

\bibitem{LC}Lee, E.H. and Cherkassky, V., 2024. Understanding Double Descent Using VC-Theoretical Framework. {\it IEEE Transactions on Neural Networks and Learning Systems}.

\bibitem{M} Mahesh, B. (2020). Machine learning algorithms-a review. {\it International Journal of Science and Research (IJSR)}, 381-386.


\bibitem{MF} Mignacco, F., Krzakala, F., Lu, Y., Urbani, P. and Zdeborova, L., (2020), The role of regularization in classification of high-dimensional noisy gaussian mixture.{\it International conference on machine learning}, 6874-6883.




\bibitem{NP} Nakkiran, P. (2019). More data can hurt for linear regression: Sample-wise double descent. {\it arXiv preprint arXiv: 1912.07242}.


\bibitem{NKBYB+} Nakkiran, P., Kaplun, G., Bansal, Y., Yang, T., Barak, B., and Sutskever, I. (2021). Deep double descent: Where bigger models and more data hurt. {\it Journal of Statistical Mechanics: Theory and Experiment}, ({\bf12}), 124003.


\bibitem{NVKM} Nakkiran, P., Venkat, P., Kakade, S. and Ma, T. (2020). Optimal regularization can mitigate double descent. {\it arXiv preprint arXiv: 2003.01897}.


\bibitem{SB} Simon, C. P. and Blume, L. (1994). {\it Mathematics for Economists}, ({\bf7}), New York: Norton.

\bibitem{SGDSB+} Spigler, S., Geiger, M., d’Ascoli, S., Sagun, L., Biroli, G. and Wyart, M. (2019). A jamming transition from under-to over-parametrization affects generalization in deep learning. {\it Journal of Physics A: Mathematical and Theoretical, 52}({\bf47}), 474001.

\bibitem{TOH2} Thrampoulidis, C., Oymak, S. and Hassibi, B. (2014). The Gaussian min-max theorem in the presence of convexity. {\it arXiv preprint arXiv: 1408.4837}.

\bibitem{TOH} Thrampoulidis, C., Oymak, S. and Hassibi, B. (2015). Regularized linear regression: A precise analysis of the estimation error. {\it Conference on Learning Theory}, 1683-1709.


\end{thebibliography}
\end{document}